%% file: main.tex
\documentclass[10pt,journal,compsoc]{IEEEtran}
\usepackage{xcolor,soul,framed} 
\usepackage{ragged2e} 
\usepackage{algorithm}
\usepackage{algpseudocode}

\usepackage[pdftex]{graphicx}
\graphicspath{{../pdf/}{../jpeg/}}
\DeclareGraphicsExtensions{.pdf,.jpeg,.png}

\usepackage[cmex10]{amsmath}
\usepackage{amsthm}
\usepackage{amssymb} 
\usepackage{thmtools, thm-restate}
\usepackage{array}
\usepackage{mdwmath}
\usepackage{bbm}
\usepackage{mdwtab}
\usepackage{eqparbox}

\usepackage{url}
\usepackage{doi} 

\usepackage{pifont} 
\usepackage[numbers,sort&compress,comma]{natbib}
\usepackage{multirow}

\usepackage{diagbox}
\usepackage{makecell}

\usepackage[english]{cleveref}
\renewcommand{\figureautorefname}{Figure}

\def\equationautorefname~#1\null{Eqn.~(#1)\null}
\def\figureautorefname~#1\null{Fig.~#1\null}
\def\lemmaautorefname~#1\null{Lemma~#1\null}

\def\CHK#1 {\textcolor{magenta}{{\bf [CHK:}~#1{\bf ]}}~}
\def\ADD#1 {\textcolor{cyan}{{\bf [ADD:}~#1{\bf}]}~}

\begin{document}
\title{MegaSR: Mining Customized Semantics and Expressive Guidance for Real-World Image Super-Resolution}

\author{
    Xinrui Li,
    Jinrong Zhang,
    Jianlong Wu, \IEEEmembership{Member,~IEEE,}
    Chong Chen,
    Liqiang Nie, \IEEEmembership{Senior Member,~IEEE,}
    Zhouchen Lin, \IEEEmembership{Fellow,~IEEE}
    
    \thanks{$\bullet$ Xinrui Li, Jinrong Zhang, Jianlong Wu, and Liqiang Nie are with the School of Computer Science and Technology, Harbin Institute of Technology, Shenzhen 518055, China (email: felix.leeovo7@gmail.com; zhangjinrong731@stu.hit.edu.cn; wujianlong@hit.edu.cn; nieliqiang@gmail.com).}
    \thanks{$\bullet$ Chong Chen is with School of Mathematical Sciences, Peking University, Beijing 100871, China (email: chenchong.cz@gmail.com).}
    \thanks{$\bullet$ Zhouchen Lin is with the State Key Lab of General AI, School of Intelligence Science and Technology, Peking University, Beijing 100871, China, also with the Institute of Artificial Intelligence, Peking University, Beijing 100871, China, and also with the Pazhou Laboratory (Huangpu), Guangzhou 510335, China (e-mail: zlin@pku.edu.cn).}
}



\IEEEtitleabstractindextext{%
\input{parts/abstract}
}

\maketitle

\ifCLASSOPTIONpeerreview
\begin{center} \bfseries EDICS Category: 3-BBND \end{center}
\fi
%


\input{parts/I_introduction}

\input{parts/II_related_work}

\input{parts/III_phenomena_analysis_and_discussion}

\input{parts/IV_methods}

\input{parts/V_experiments}

\input{parts/VI_conclusion}

\ifCLASSOPTIONcaptionsoff
  \newpage
\fi





\bibliographystyle{IEEEtran}
\bibliography{IEEEabrv,Bibliography}


\end{document}

%% file: parts/abstract.tex
\begin{abstract}
\justifying
Text-to-image (T2I) models have ushered in a new era of real-world image super-resolution (Real-ISR) due to their rich internal implicit knowledge for multimodal learning.
Although bringing high-level semantic priors and dense pixel guidance have led to advances in reconstruction, we identified several critical phenomena by analyzing the behavior of existing T2I-based Real-ISR methods:
(1) Fine detail deficiency, which ultimately leads to incorrect reconstruction in local regions.
(2) Block-wise semantic inconsistency, which results in distracted semantic interpretations across U-Net blocks.
(3) Edge ambiguity, which causes noticeable structural degradation.
Building upon these observations, we first introduce MegaSR, which enhances the T2I-based Real-ISR models with fine-grained customized semantics and expressive guidance to unlock semantically rich and structurally consistent reconstruction.
Then, we propose the Customized Semantics Module (CSM) to supplement fine-grained semantics from the image modality and regulate the semantic fusion between multi-level knowledge to realize customization for different U-Net blocks.
Besides the semantic adaptation, we identify expressive multimodal signals through pair-wise comparisons and introduce the Multimodal Signal Fusion Module (MSFM) to aggregate them for structurally consistent reconstruction.
Extensive experiments on real-world and synthetic datasets demonstrate the superiority of the method.
Notably, it not only achieves state-of-the-art performance on quality-driven metrics but also remains competitive on fidelity-focused metrics, striking a balance between perceptual realism and faithful content reconstruction.
\end{abstract}

%% file: parts/I_introduction.tex
\section{Introduction}
\label{sec:intro}

\begin{figure*}[t]
    \centering
    \includegraphics[width=0.98\linewidth]{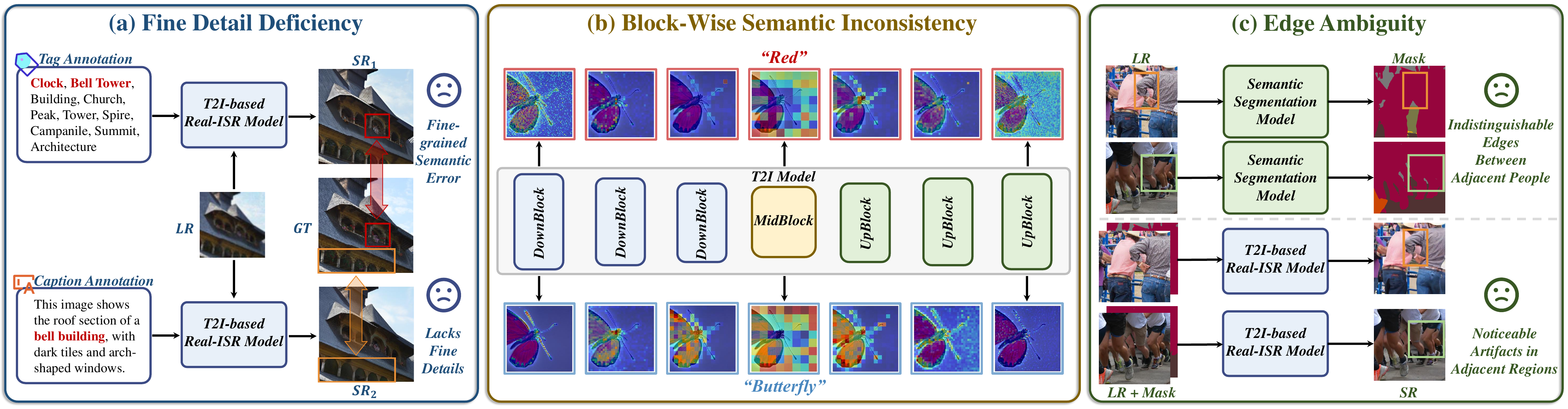}
    \vspace{-1em}
    \caption{
    Different phenomena observed in existing methods.
    (a) Solely using textual semantics for reconstruction results in erroneous fine-grained semantics or missing fine details.
    (b) The blocks at the two ends are sensitive to attribute-level concepts, while the middle blocks focus on instance-level concepts.
    (c) Semantic segmentation masks introduce edge ambiguity within the same semantic region and lead to artifacts in the results.
    }
    \vspace{-1em}
    \label{fig:intro}
\end{figure*}

\IEEEPARstart{R}{eal-world} image super-resolution (Real-ISR) aims to reconstruct high-resolution (HR) images from low-resolution (LR) counterparts captured in real scenarios under realistic imaging constraints, which not only requires filling in global pixel regions but also demands plausibly recovering the missing fine details.
Traditional image super-resolution (ISR) methods~\cite{dong_srcnn_2016, zhong_dualsr_2025, chen_hat_2025} that rely on ideal bicubic downsampling kernels struggle to address the unstructured pixel loss and diverse blur patterns in real-world scenarios, resulting in limited applicability.
To overcome these limitations, recent researches~\cite{miaoyu_recdiffsr_2024, tong_stidiffsr_2024} have adapted text-to-image (T2I) models~\cite{alexander_ddpm_2021, william_dit_2023} to handle complex pixel degradation and generate perceptually realistic details, where the rich prior knowledge embedded in these models can effectively enhance visual quality.
However, the T2I frameworks are primarily designed for image generation and still deviate from the objectives of Real-ISR, for which systematic analyses remain limited.

Existing T2I methods generate images by sampling from a feature space that is enriched with prior knowledge under specific control signals~\cite{alexander_ddpm_2021, william_dit_2023}.
Although they have been successfully adapted from standard image generation tasks to Real-ISR tasks by modifying multimodal input~\cite{wang_stablesr_2024, yang_pasd_2024}, the generation architectures dominated by high-level semantic priors still exhibit a discrepancy with the low-level pixel alignment required for super-resolution.
Real-ISR requires that the sampled outputs not only align with the semantic signals but also achieve pixel-level coherence with the LR images, where the constraints are more stringent than those in conventional image generation tasks.
Existing T2I-based Real-ISR methods~\cite{wu_seesr_2024, tsao_holisdip_2024} often focus on supplementing richer prior knowledge for sampling in the feature space, while overlooking strict control over the sampling process itself.
Specifically, we analyze the core mismatch of existing T2I-based Real-ISR frameworks and summarize several critical phenomena as follows: 
(1) \textit{Fine detail deficiency.}
Following conventional image generation tasks, current methods~\cite{yang_pasd_2024, wu_seesr_2024} incorporate text descriptions to assist in the reconstruction of pixel details.
Unlike the easily described objects in the image generation scenario, the absent patterns in LR images for Real-ISR often consist of fine texture details, posing a challenge for only textual semantic cues.
As shown in \autoref{fig:intro}(a), the reconstructed images either reconstruct flowers as bell-shaped objects following the descriptions or omit the non-salient objects.
(2) \textit{Block-wise semantic inconsistency.}
Conventional T2I frameworks have not fully explored the fine-grained semantic control within the U-Net, which is sufficient for generating regular images but remains inadequate for Real-ISR tasks that require pixel-level alignment.
We observed that different U-Net~\cite{ronneberger_unet_2015} blocks have specific preferences to perceive distinct levels of semantics.
As shown in \autoref{fig:intro}(b), the blocks at the two ends exhibit stronger responses to the attribute-level concepts (e.g., ``red'' or ``rough''), whereas the middle blocks are more sensitive to the instance-level concepts (e.g., ``butterfly'' or ``stone'').
However, most existing methods feed the same textual description to all blocks within the U-Net, distracting the model from attending to such block-specific semantic preferences.
(3) \textit{Edge ambiguity.}
Although sematic segmentation masks~\cite{cheng_mask2former_2022, li_maskdino_2023} are widely used as pixel-level supervisory signals~\cite{tsao_holisdip_2024}, they introduce an ambiguity in delineating boundaries when identical semantic objects are spatially adjacent, as shown in \autoref{fig:intro}(c).
Feeding these masks into T2I-based Real-ISR models~\cite{tsao_holisdip_2024} introduces noticeable artifacts in adjacent regions, which has a marginal impact on conventional image generation tasks, but degrades the performance in Real-ISR tasks.
We will present detailed evidence in \autoref{sec:phe}.

To address the issues mentioned above, we introduce \textbf{MegaSR}, which \textbf{m}ines fine-grained customized s\textbf{e}mantics and expressive \textbf{g}uid\textbf{a}nce for Real-I\textbf{SR}.
Based on T2I U-Net~\cite{rombach_ldm_2022}, the proposed method takes LR images as input.
On the one hand, it leverages RAM~\cite{zhang_ram_2024} and prior-guided fine-tuned CLIP~\cite{radford_clip_2021} vision encoder to extract coarse-grained textual semantics and fine-grained visual semantics, and dynamically adjust their weights across different U-Net blocks.
On the other hand, it employs prior-guided fine-tuned signal extractors to derive multimodal guidance signals, which are progressively injected into the intermediate representations. 
Ultimately, it produces HR images that are both semantically rich and structurally consistent.
It consists of the following improvements:

Building upon the shared text-image embedding spaces learned by T2I models on billions of data, we introduce the Dual-Path Cross-Attention (DPCA) mechanism to enable interactions between textual and visual semantics.
Specifically, DPCA comprises two parallel branches: one branch retains the original cross-attention mechanism~\cite{rombach_ldm_2022} over the text modality, while the other branch leverages extracted image representations to perform complementary cross-attention.
The hidden states derived from the two branches are subsequently fused to form a unified representation, providing an enriched multi-granularity context for subsequent stages.

To satisfy the distinct semantic requirements of different U-Net blocks, we present a Learnable Gated Weight Adaptation Module (LGWAM), which dynamically regulates the relative ratio of the textual and visual branches in DPCA.
As concluded from \autoref{fig:intro}(b) and \autoref{sec:phe}, different blocks focus on distinct levels of semantics.
LGWAM achieves this multi-level balance by adaptively scaling the hidden states of the visual branch.
On the one hand, it facilitates semantic coordination across different branches.
On the other hand, it preserves the capabilities of the T2I models and simplifies the training process.
Together with DPCA, LGWAM constitutes the Customized Semantics Module (CSM), integrating text–image semantic fusion with adaptive weighting.

Beyond the semantic adaptation, to address the ambiguity inherent in semantic segmentation masks, we investigate multi-modal guidance and propose a Multimodal Signal Fusion Module (MSFM) to inject expressive and non-redundant signals into the T2I backbone.
Specifically, we conducted pair-wise comparative experiments in \autoref{sec:phe} and ultimately identified HED boundaries~\cite{xie_hed_2017}, depth maps~\cite{yang_depthanythingv2_2024}, and semantic segmentation masks~\cite{li_maskdino_2023} as the most effective modalities.
HED boundaries are sparse edge signals that enable precise localization of structures and shapes within regions sharing the same semantics.
Depth maps and segmentation masks are dense pixel-level signals, providing multi-dimensional pattern cues for spatial reasoning and contextual guidance.
Hence, we design a two-stage fusion strategy within the MSFM, where depth maps and segmentation masks are fused in the first stage, and then collaborate with HED boundaries in the second stage to exert precise control over the diffusion process.

To sum up, our contributions are four-fold:
\begin{itemize}
    \item We summarize the limitations of existing T2I-based Real-ISR methods through comprehensive analysis, providing insights for further exploration.

    \item We introduce MegaSR, which addresses the deficiency, inconsistency, and ambiguity of T2I-based Real-ISR methods and enhances the semantic richness and structural consistency of the reconstruction.

    \item We present two specialized modules for context-aware and structure-aware Real-ISR:
    (1) Customized Semantics Module, which supplements fine-grained semantics and customizes the multi-level knowledge to enhance semantic adaptation.
    (2) Multimodal Signal Fusion Module, which aggregates expressive multimodal signals to exert pixel-level guidance.
    
    \item Comprehensive experiments on real-world and synthetic datasets demonstrate the effectiveness of MegaSR.
    Notably, it strikes a balance on quality-driven metrics and fidelity-focused metrics, highlighting the overall robustness.
\end{itemize}

%% file: parts/II_related_work.tex
\section{Related Work}
\label{sec:related_work}
\subsection{Mapping-based Image Super-Resolution}
\label{subsec:isr}
Mapping-based ISR methods employ deep neural networks~\cite{zaremba_rnn_2014, vaswani_transformer_2017} to directly model the projection from LR images to their HR counterparts.
Such LR inputs are obtained by straightforward downsampling of the HR images.
And the parameters are optimized in an end-to-end manner using fidelity-focused objectives.

Existing ISR methods can be categorized into three main types based on the feature extraction block: CNN-based methods, transformer-based methods, and hybrid methods.
Each type presents specific limitations for image quality.
(1) \textit{CNN-based methods}~\cite{kim_vdsr_2016, kim_drcn_2016, tai_drrn_2017}, since the pioneering work of SRCNN~\cite{dong_srcnn_2016}, have been extensively studied with more complicated architectures, such as Laplacian pyramids~\cite{lai_fastlapsr_2018} and U-shaped designs~\cite{cheng_usr1_2019}.
While these methods offer effective image representations, long-range dependencies are often compromised by the constrained convolution kernel size.
(2) \textit{Transformer-based methods}~\cite{zhang_rcan_2018, chen_ipt_2021, liang_swinir_2021} integrate the attention mechanism to model long-range dependencies.
However, their quadratic computational cost hinders efficiency and practical implementation.
(3) \textit{Hybrid methods}~\cite{chen_hat_2025, chen_dat_2023, liu_catanet_2025} attempt to strike a balance between local feature extraction and global context modeling by combining CNNs and transformers.
However, fidelity-focused objectives dominate the reconstruction process, resulting in blurred results.

To overcome these bottlenecks, this study aims to improve the perceptual quality with the T2I backbone.

\subsection{GAN-based Image Super-Resolution}
\label{subsec:gan_isr}
GAN-based~\cite{goodfellow_gan_2014} ISR methods~\cite{ledig_srgan_2017, hasan_pcsrgan_2025} consist of two components: a generator and a discriminator.
In addition, they incorporate perceptual loss~\cite{johnson_ploss_2016} and adversarial loss to enhance the realism of the reconstructed images.
In terms of data synthesis, previous methods incorporate a high-order degradation pipeline~\cite{wang_realesrgan_2021} to generate LR-HR pairs and simulate real-world degradation patterns, thus enhancing the robustness of the methods.

The generator is designed to synthesize reconstructed outputs that closely resemble real samples and deceive the discriminator.
Hence, existing methods~\cite{ledig_srgan_2017, zhongxi_rethinkinggan_2023} initialize the generator with the deep neural network architectures introduced in \autoref{subsec:isr} to ensure representational capacity.
The discriminator, by contrast, is introduced to distinguish the reconstructed outputs from real samples, compelling the generator to progressively model the real distribution with higher fidelity.
Existing discriminator architectures include VGG~\cite{simonyan_vgg_2015}, which provides a holistic assessment, and U-Net~\cite{ronneberger_unet_2015}, which delivers pixel-level feedback.
Although GAN-based ISR methods improve the perceptual quality of reconstructed results, they inherently suffer from training instability~\cite{miyato_ganinsta_2018}, restricting their practical application.

To this end, this study focuses on T2I models that possess stronger prior knowledge and improved training stability.

\subsection{T2I-based Image Super-Resolution}
\label{subsec:t2i_isr}

T2I-based ISR methods typically comprise an autoencoder, a denoising backbone, and a controlnet mechanism.
Existing ISR refinements to the T2I framework can be categorized based on their main contributions into LR-focused, text-focused, and dense signal-focused.

\textit{LR-focused refinements} leverage the inherent LR inputs to directly control the diffusion process through projection, preprocessing, and alignment.
These methods rely on robust representation modeling modules, such as time-aware encoders~\cite{wang_stablesr_2024}, deep neural networks~\cite{lin_diffbir_2024}, and alignment modules~\cite{chen_faithdiff_2025}, to fully exploit the signals in the LR inputs for precise control.

\textit{Text-focused refinements} utilize pre-trained caption or tagging models to produce visually grounded annotations that serve as a global catalyst during diffusion~\cite{yang_pasd_2024, wu_seesr_2024}.
However, their quality heavily depends on high-level models~\cite{he_resnet_2016, li_blip_2022}, leading to issues such as the lack of fine-grained components and potential distractions.
For instance, the tags extracted by RAM~\cite{zhang_ram_2024} cause SeeSR~\cite{wu_seesr_2024} to reconstruct distorted or erroneous details, significantly degrading the visual quality of the reconstruction.

\textit{Dense signal-focused refinements} emphasize pixel-level signals for Real-ISR.
For example, HoliSDiP~\cite{tsao_holisdip_2024} integrates CLIP~\cite{radford_clip_2021} to embed semantic segmentation masks for dense semantic guidance, while SegSR~\cite{xiao_segsr_2024} employs a parallel diffusion framework for semantic segmentation masks and RGB images.
Nevertheless, these approaches, while increasing computational overhead, rely solely on single-modal signals to guide the T2I process, causing ambiguity in delineating the contours of entities sharing the same semantics.

In this study, we introduce a customized semantics module for enhanced semantic awareness, as well as a multi-modal signal fusion module to ensure complementary and consistent feature representation across modalities.

%% file: parts/III_phenomena_analysis_and_discussion.tex
\section{Preliminaries and Problem Statement}
\label{sec:phe}

\begin{figure*}[htbp]
    \centering
    \includegraphics[width=0.98\linewidth]{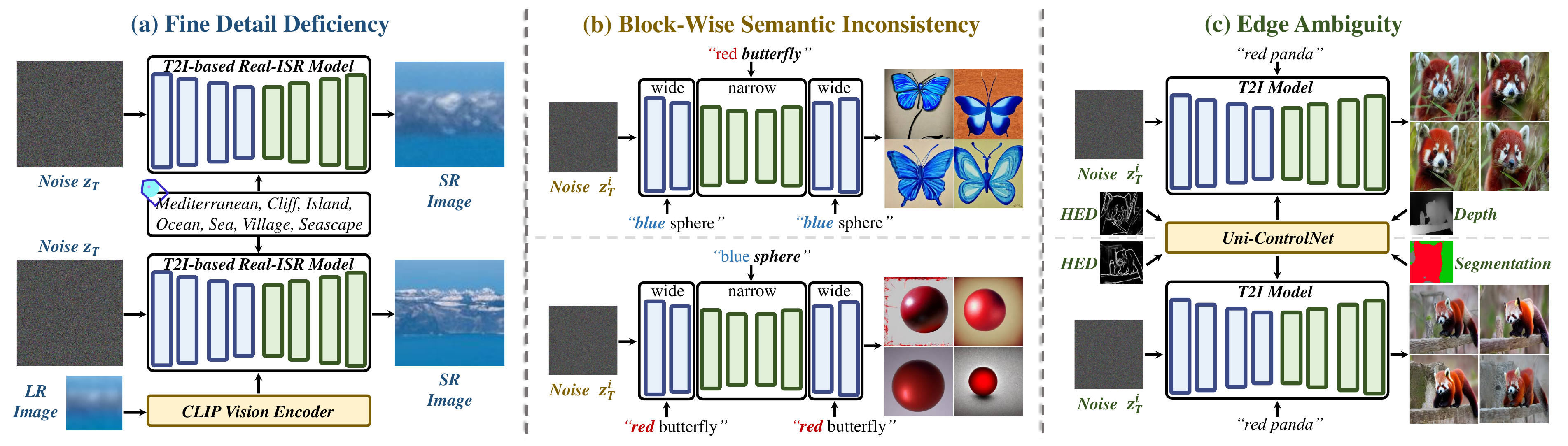}
    \vspace{-1em}
    \caption{
    Detailed statements of the phenomena.
    (a) Incorporating fine-grained visual semantics contributes to both improved visual clarity and enhanced semantic fidelity.
    (b) Applying different prompts to U-Net blocks at varying widths demonstrates that wide and narrow blocks in T2I models play distinct roles.
    (c) Determining the relative intensity between signals by pairing them as inputs to Uni-ControlNet~\cite{zhao_unicontrolnet_2023} and evaluating the structural differences in the generated results.
    }
    \vspace{-1em}
    \label{fig:motivation}
\end{figure*}

In this section, we provide more detailed statements of the phenomena introduced in \autoref{sec:intro}.
Specifically, we first present a further analysis of the fine detail deficiency in \autoref{subsec:fine_grained}.
Then, we conduct qualitative and quantitative experiments to demonstrate semantic inconsistency across the U-Net blocks in \autoref{subsec:semantics}.
Next, to explore multimodal signals to address edge ambiguity, we compare multimodal guidance signals and identify the most expressive and non-redundant modalities in \autoref{subsec:signals}.

\subsection{Fine Detail Deficiency}
\label{subsec:fine_grained}
To verify the effectiveness of fine-grained visual semantics, we extracted embeddings that preserve non-salient semantic cues from the LR image and injected them into the T2I-based Real-ISR model.
As illustrated in \autoref{fig:motivation}(a), when only coarse-grained textual semantics are available, the central object appears to be overly smoothed and indistinguishable.
In contrast, supplementing the model with fine-grained visual semantics leads to clearer and semantically more accurate results.
These observations demonstrate that textual descriptions extracted from degraded LR images through existing annotation models are insufficient to capture subtle image details and that the integration of fine-grained visual semantics effectively compensates for this limitation.
Quantitative evaluations are presented in \autoref{subsubsec:ablation_csm}.

\subsection{Block-Wise Semantic Inconsistency}
\label{subsec:semantics}
In \autoref{sec:intro}, we observe that there is a semantic inconsistency across different U-Net blocks.
Considering their architectural characteristics, the primary distinction is derived from the variation in the effective receptive field caused by the downsampling operations.
Hence, an intuitive hypothesis is that blocks of different widths perceive distinct levels of semantics: wider blocks with larger receptive fields capture low-level attributes (e.g., color, texture), whereas narrower ones with limited receptive fields focus on high-level semantics (e.g., category, instance).
To validate this, we first conducted a qualitative experiment and then extended it to a large-scale quantitative evaluation.

As shown in \autoref{fig:motivation}(b), we paired the prompts \textit{``red butterfly''} and \textit{``blue sphere''}, which differ in both color and object descriptions, into the narrow and wide blocks of the T2I model~\cite{rombach_ldm_2022}, respectively.
Then we swapped them.
In the first combination, the generated images depicted a blue butterfly.
After the inputs were exchanged, the results became a red sphere.
This indicates that the semantics perceived by the wide block are closely related to low-level attributes in the output, whereas the semantics perceived by the narrow block are more aligned with high-level concepts.

To provide quantitative evidence, we designed an automated image generation and evaluation pipeline.
We first collected 1,000 pairs of prompts generated by a Large Language Model (LLM)~\cite{yang_qwen3_2025}.
\begin{equation}
    \mathcal{P}^1, \mathcal{P}^2 = \Psi(\mathcal{T}),
\end{equation}
where $\mathcal{P}^1, \mathcal{P}^2$ are prompt pairs, $\Psi$ is the LLM and $\mathcal{T}$ is the template for prompt generation.
Then each pair of prompts was input into U-Net blocks at different widths of the T2I model~\cite{rombach_ldm_2022} to generate a set of images.
\begin{equation}
    \left\{ \mathcal{I}^i \right\}_{i = 1}^{n = 4} = \Theta(\mathcal{P}^1, \mathcal{P}^2, n),
\end{equation}
where $\Theta$ is the T2I model and $\mathcal{I}^i$ is the generated image.
Finally, we utilized a Multimodal Large Language Model (MLLM)~\cite{bai_qwen25vl_2025} to evaluate the prompt with which each image was more aligned in terms of low-level attributes and high-level semantics.
\begin{equation}
    \begin{aligned}
        \mathcal{C}_{low} &= \Phi(\mathcal{I}^i), \mathcal{C}_{low} \in \left\{ \mathcal{P}^1, \mathcal{P}^2 \right\}, \\
        \mathcal{C}_{high} &= \Phi(\mathcal{I}^i), \mathcal{C}_{high} \in \left\{ \mathcal{P}^1, \mathcal{P}^2 \right\},
    \end{aligned}
\end{equation}
where $\Phi$ is the MLLM and $\mathcal{C}_{low}, \mathcal{C}_{high}$ are the evaluation results.
For each pair of prompts, we conducted two experimental setups.
In one, $\mathcal{P}^1$ was fed into the wide blocks and $\mathcal{P}^2$ into the narrow blocks; in the other, they were swapped.

The results are shown in \autoref{tab:semantics}.
When $\mathcal{P}^1$ and $\mathcal{P}^2$ were fed into wide and shallow blocks, respectively, the combination $\mathcal{P}^1 + \mathcal{P}^2$ achieves the highest accuracy.
However, after swapping, $\mathcal{P}^2 + \mathcal{P}^1$ performs best.
This quantitative evaluation supports the hypothesis that wide and narrow blocks in the T2I models play distinct roles in terms of semantic perception.
However, uniform semantic inputs in existing methods limit the ability of different U-Net blocks to concentrate on expressing task-relevant cues.

\input{tables/motivation_semantics}

\subsection{Edge Ambiguity}
\label{subsec:signals}

The edge ambiguity observed in semantic segmentation masks arises from the limited knowledge available in unimodal signals.
To mitigate this, we consider incorporating multimodal guidance signals to enhance the information density.
However, a question arises: among the diverse guidance signals, which ones are the most expressive while avoiding redundancy?
To investigate this, we first design a preliminary experiment to intuitively compare the intensity of different modalities.
Furthermore, to improve generalization, we also introduced an automated data generation and evaluation pipeline, which facilitates large-scale experiments for a more systematic validation of the observation.

\input{tables/motivation_signals}

As shown in \autoref{fig:motivation}(c), we paired guidance signals with identical semantics but contradictory structures for Uni-ControlNet~\cite{zhao_unicontrolnet_2023} and evaluated their relative intensity.
In the first group, we utilized HED boundaries and depth maps as guidance signals.
Both describe the \textit{``red panda''}, but they differ in the shape of the main object.
The generated images exhibit more consistency with the HED boundaries.
Hence, we consider the HED boundaries to be more salient than the depth maps.
In the second group, we replaced the signals with HED boundaries and semantic segmentation masks.
Compared with semantic segmentation masks, the HED boundaries also exerted a stronger influence.

To systematically quantify the intensity of different guidance signals, we designed an automated data generation and evaluation pipeline to compare HED boundaries~\cite{xie_hed_2017}, Canny edges, sketch maps, depth maps~\cite{yang_depthanythingv2_2024}, semantic segmentation masks~\cite{li_maskdino_2023}, and pose maps~\cite{lvmin_controlnet_2023}.
We collected 1,000 paired prompts generated by the LLM~\cite{yang_qwen3_2025}.
\begin{equation}
    \mathcal{P}_1, \mathcal{P}_2 = \Psi(\mathcal{T}'),
\end{equation}
where $\mathcal{P}_1, \mathcal{P}_2$ are prompt pairs and $\mathcal{T}'$ is the template for prompt generation.
Each pair describes the same object but with different structures.
Then, we utilized the prompts to generate image pairs $\mathcal{I}_1$ and $\mathcal{I}_2$ with the T2I model~\cite{rombach_ldm_2022}.
\begin{equation}
    \mathcal{I}_1, \mathcal{I}_2 = \Theta(\mathcal{P}_1), \Theta(\mathcal{P}_2).
\end{equation}
Following the preliminary experiment, we extracted guidance signals from image pairs with a set of extractors and encoded them as conditional inputs for the controllable image generation unimodal model~\cite{zhao_unicontrolnet_2023}.
\begin{equation}
    \begin{aligned}
        \left\{ \mathcal{G}^1_i \right\}_{i = 1}^{n = 6}, \left\{ \mathcal{G}^2_i \right\}_{i = 1}^{n = 6} &= \eta(\mathcal{I}_1), \eta(\mathcal{I}_2), \\
        \left\{ \hat{\mathcal{I}_i} \right\}_{i = 1}^{n = 4} &= \Omega(\left\{ \mathcal{G}^1_i \right\}_{i = 1}^{n = 6}, \left\{ \mathcal{G}^2_i \right\}_{i = 1}^{n = 6}), \\
    \end{aligned}
\end{equation}
where $\mathcal{G}_i$ are the guidance signals, $\eta$ is the set of extractors, $\hat{\mathcal{I}_i}$ is the generated image conditioned on the extracted signals and $\Omega$ is the controllable image generation model.
After generation, we evaluated the similarity $\mathcal{C}_{sim}$ utilizing the MLLM~\cite{bai_qwen25vl_2025} and aggregated the final statistics.
\begin{equation}
    \mathcal{C}_{sim} = \Phi(\hat{\mathcal{I}_i}, \mathcal{I}_1, \mathcal{I}_2), \mathcal{C}_{sim} \in \left\{ \mathcal{I}_1, \mathcal{I}_2 \right\}.
\end{equation}

The results are shown in \autoref{tab:signals}.
For each pair of signals, taking (HED, Depth) as an example, if the generated image is closer to HED boundaries, we increment the count for (HED, Depth).
Conversely, if the image aligns with the depth map, we increment the count for (Depth, HED).
To this end, the sum of each row represents the frequency of the signal that is favored over others, while the sum of each column represents the frequency of the signal that is outperformed.
Hence, HED boundaries emerge as the most expressive signal, followed by Canny edges, sketch maps, depth maps, segmentation masks, and pose maps.
Since HED boundaries, Canny edges, and sketch maps belong to the category of edge-based signals, we retain HED boundaries as a representative choice.
Additionally, depth maps and segmentation masks are preserved, as they capture complementary spatial knowledge, including object positioning along the vertical axis and object contexts along the horizontal axis.
Finally, pose maps are excluded since pose information is often unavailable in the Real-ISR tasks.

%% file: tables/motivation_semantics.tex
\begin{table}[t]
    \renewcommand\arraystretch{1.2}
    \caption{
    The quantitative evaluation results of width-specific semantic requirements in T2I models.
    }
    \label{tab:semantics}
    \centering
    \resizebox{\linewidth}{!}{
        \begin{tabular}{cccccc}
            \Xhline{1.25pt}
            $\mathcal{C}_{low} + \mathcal{C}_{high}$ & $\mathcal{P}^1 + \mathcal{P}^1$ & $\mathcal{P}^1 + \mathcal{P}^2$ & $\mathcal{P}^2 + \mathcal{P}^1$ & $\mathcal{P}^2 + \mathcal{P}^2$ & others \\
            \Xhline{0.75pt}
            Bef. Acc. & 7.9\% & \textbf{71.5\%} & 0.4\% & 20.0\% & 0.2\% \\
            Aft. Acc. & 29.5\% & 0.05\% & \textbf{65.3\%} & 5.1\% & 0.05\% \\
            \Xhline{1.25pt}
        \end{tabular}
    }
    \vspace{-1em}
\end{table}

%% file: tables/motivation_signals.tex
\begin{table}[t]
    \renewcommand\arraystretch{1.3}
    \vspace{-0.2em}
    \caption{
    The quantitative evaluation results of the relative intensity between different guidance signals.
    }
    \label{tab:signals}
    \centering
    \resizebox{\linewidth}{!}{
        \begin{tabular}{ccccccc|cc}
            \Xhline{1.25pt}
             & HED & Canny & Sketch & Depth & Seg & Pose & Success & Rank \\
            \Xhline{0.75pt}
            HED & - & 1286 & 1449 & 1472 & 1487 & 1486 & 7180 & \textbf{\textit{1}} \\
            Canny & 714 & - & 1335 & 1370 & 1432 & 1432 & 6283 & \textbf{\textit{2}} \\
            Sketch & 551 & 665 & - & 1078 & 1246 & 1263 & 4803 & \textbf{\textit{3}} \\
            Depth & 528 & 630 & 922 & - & 1232 & 1253 & 4565 & \textbf{\textit{4}} \\
            Seg & 513 & 568 & 754 & 768 & - & 1111 & 3714 & \textbf{\textit{5}} \\
            Pose & 514 & 568 & 737 & 747 & 889 & - & 3455 & \textbf{\textit{6}} \\
            \Xhline{0.75pt}
            Failure & 2820 & 3717 & 5197 & 5435 & 6286 & 6545 & - & - \\
            Rank & \textbf{\textit{1}} & \textbf{\textit{2}} & \textbf{\textit{3}} & \textbf{\textit{4}} & \textbf{\textit{5}} & \textbf{\textit{6}} & - & - \\
            \Xhline{1.25pt}
        \end{tabular}
    }
    \vspace{-0.7em}
\end{table}

%% file: parts/IV_methods.tex
\section{Methodology}
\label{sec:method}
In this section, we introduce our method in detail.
We first describe the overall framework of our proposed method.
Next, we elaborate on each component of the proposed method:
(a) Prior-Guided Fine-Tuning Strategies (\autoref{subsec:module0}),
(b) Customized Semantics Module, consisting of the Dual-Path Cross-Attention mechanism and the Learnable Gated Weight Adaptation Module (\autoref{subsec:module1}), and
(c) Multimodal Signal Fusion Module (\autoref{subsec:module2}).

\begin{figure*}[htbp]
    \centering
    \includegraphics[width=0.98\linewidth]{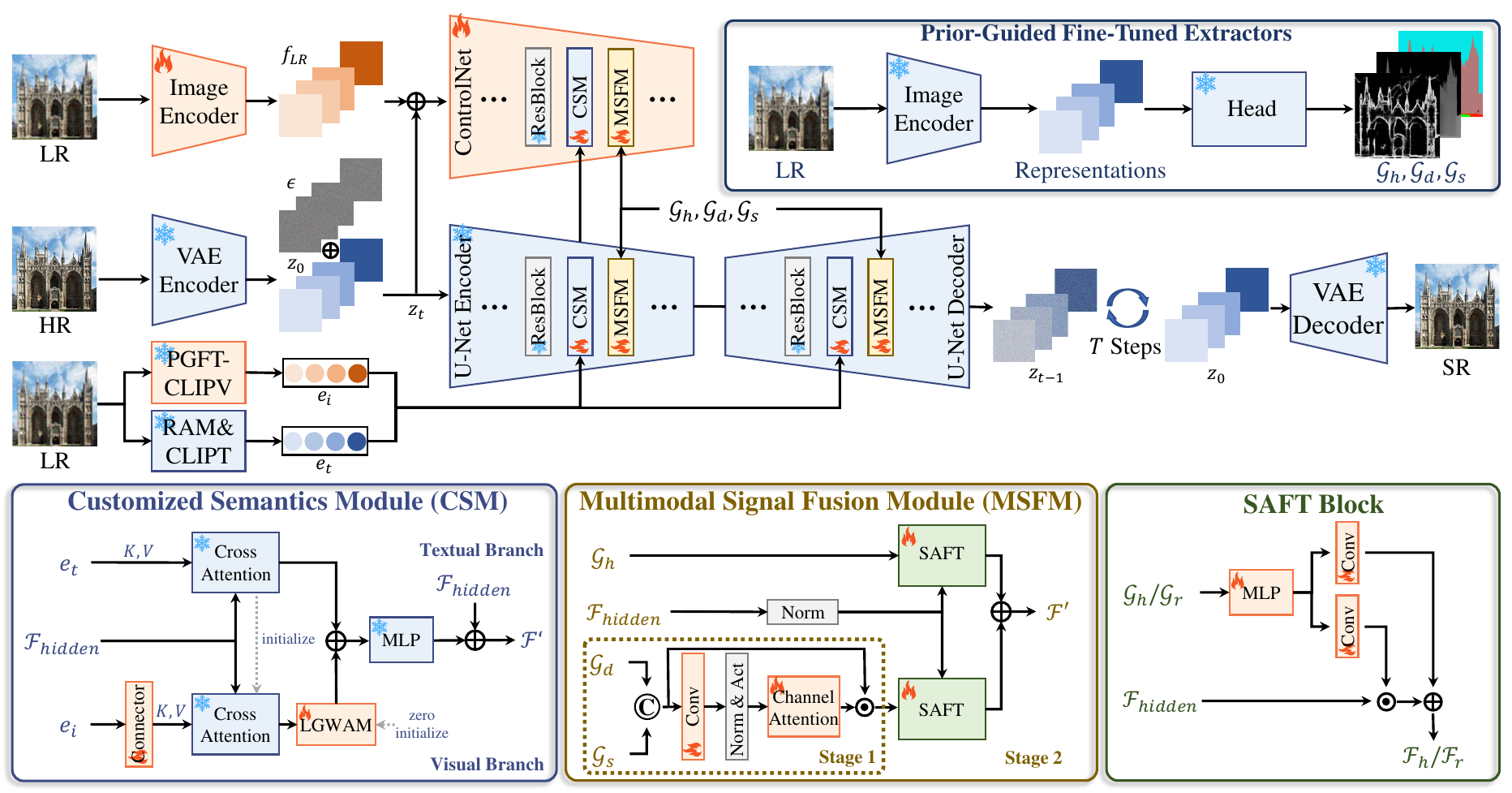}
    \vspace{-1em}
    \caption{
    Framework of the proposed method.
    Firstly, based on the T2I U-Net, it takes LR images as input.
    Then, it utilizes RAM~\cite{zhang_ram_2024} and PGFT-CLIPV model to extract coarse-grained textual and fine-grained visual semantics to DPCA, and dynamically adjusts their weights at different U-Net blocks with LGWAM.
    Next, it employs prior-guided fine-tuned extractors to obtain multimodal signals, which are progressively injected into the representations of T2I models via MSFM. 
    Finally, it produces HR images that are both semantically rich and structurally consistent.
    }
    \vspace{-0.5em}
    \label{fig:framework}
\end{figure*}

\subsection{Framework Overview}
\label{subsec:framework_overview}

The overall framework of the proposed MegaSR is shown in \autoref{fig:framework}.
Unlike existing Real-ISR methods that rely solely on textual semantics and overlook block-specific semantic requirements while incorporating unimodal pixel-level signals for guidance, MegaSR delivers width-specific fine-grained semantics and multimodal signals to enhance semantic richness and structural consistency.

Typically, signal extractors are designed to handle clear inputs.
Hence, in the Real-ISR tasks, a fine-tuning process is essential to adapt them to LR inputs.
However, some extractors inherit the degradation priors to some extent due to the low-level data augmentations during pre-training.
Based on the extent of these priors, we design two fine-tuning strategies to enhance their awareness of degradation, which will be introduced in \autoref{subsec:module0}.

Leveraging the multimodal alignment capabilities of the T2I models, we design the Dual-Path Cross-Attention (DPCA) mechanism to enrich the fine-grained semantics.
It encodes images into the same high-dimensional embedding space as textual tags to participate in attention interaction.
Meanwhile, to meet the requirements of different U-Net blocks for distinct levels of semantics, we present LGWAM to adjust the weights of the visual branch in DPCA, allowing fine-grained semantic control with minimal influence on the capabilities of the T2I model.
Together, DPCA and LGWAM constitute the Customized Semantics Module (CSM), which will be described in \autoref{subsec:module1}.

For multimodal signal fusion, we introduce the MSFM to progressively inject HED boundaries, depth maps, and semantic segmentation masks into the T2I representations.
Specifically, we first derive these signals with the prior-guided fine-tuned extractors.
Considering the characteristics of different signals, we design a two-stage fusion pipeline within the DPCA.
In the first fusion stage, we aggregate depth maps and semantic segmentation masks along the channel dimension to derive multi-dimensional signals.
In the second modulation stage, the multi-dimensional signals and HED boundaries are separately injected to modulate the latent representations of the T2I models.
More details are presented in \autoref{subsec:module2}.

\subsection{Prior-Guided Fine-Tuning Strategies}
\label{subsec:module0}

To adapt to LR inputs, we adopt pre-trained signal extractors as base models and apply prior-guided fine-tuning (PGFT) strategies to enhance their degradation awareness.
However, we empirically found that some pre-trained signal extractors already possess an inherent ability to partially handle degraded inputs.
Therefore, we design two distinct fine-tuning strategies based on the degradation priors intrinsically embedded in them.

\begin{figure*}[htbp]
    \centering
    \includegraphics[width=0.98\linewidth]{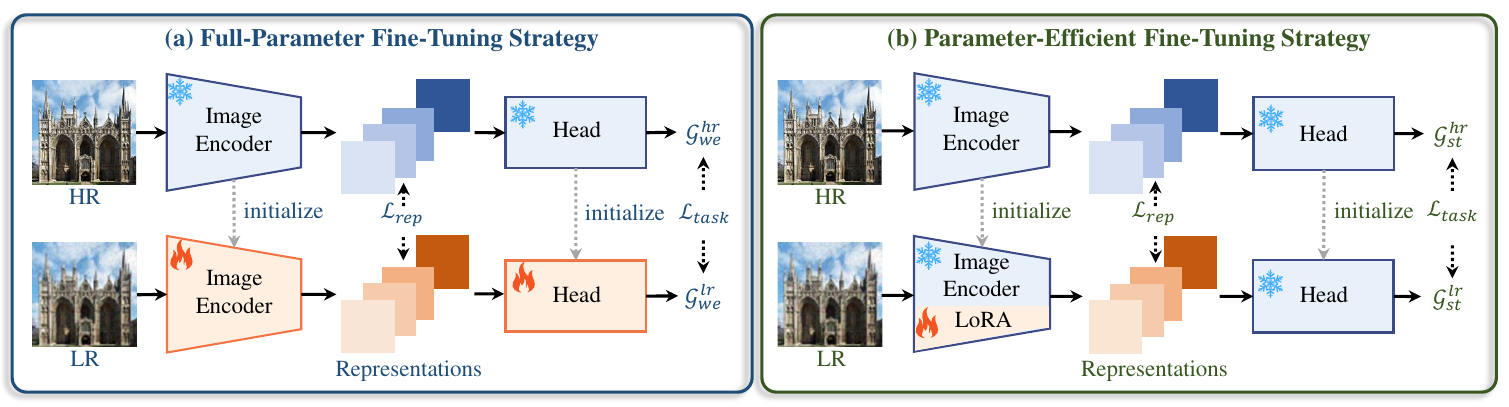}
    \vspace{-1em}
    \caption{
    Prior-guided fine-tuning strategies of the signal extractors.
    (a) For extractors with weaker degradation priors, we apply full-parameter fine-tuning to ensure flexibility for adaptation.
    (b) For extractors with stronger priors, we accelerate the process using parameter-efficient fine-tuning.
    }
    \vspace{-1em}
    \label{fig:dages}
\end{figure*}

For extractors with weaker degradation priors, we design a full-parameter fine-tuning strategy to provide greater flexibility for adapting them to LR inputs.
As shown in \autoref{fig:dages}(a), the HR image is passed through a frozen extractor to extract the image representations and guidance signals.
Similarly, the LR image goes through the same process, except that both the backbone and the head are trainable and initialized with those of the HR counterparts.

In contrast, to preserve the capabilities and enhance the fine details, we adopt a parameter-efficient fine-tuning strategy for extractors with stronger priors.
As illustrated in \autoref{fig:dages}(b), LoRA (r=8)~\cite{hu_lora_2022} is integrated into the backbone while all other parameters are kept frozen.

In both settings, we force both the representations and the signals from the LR branch to closely align with those of the HR branch to improve the alignment capability and robustness~\cite{wu_seesr_2024}.
With effective fine-tuning, the proposed method incorporates more precise pixel-level guidance signals, which facilitates improved reconstruction quality.

\subsection{Customized Semantics Module}
\label{subsec:module1}

The Customized Semantics Module (CSM) is proposed to supplement fine-grained image semantics and enable dynamic adjustment of multi-level semantics across U-Net blocks of varying widths.
As shown in \autoref{fig:framework}, it comprises two components: the Dual-Path Cross-Attention (DPCA) mechanism and the Learnable Gated Weight Adaptation Module (LGWAM).

Specifically, the DPCA mechanism consists of two branches: one frozen branch for coarse-grained textual semantics (textual branch) and one trainable branch for fine-grained visual semantics (visual branch).
The textual branch directly utilizes the cross-attention mechanism of the original T2I models.
For the visual branch, we first employ the prior-guided fine-tuned CLIP vision encoder~\cite{radford_clip_2021} (PGFT-CLIPV) to extract image embeddings from LR images.
These embeddings are passed through a linear connector to enable space transformation.
Next, the transformed embeddings are processed by a frozen cross-attention mechanism, which is initialized with that of the textual branch to facilitate interaction with the latent representations.
The whole process is described as follows.
\begin{equation}
    \begin{aligned}
        e_t &= \text{CLIPT}(\mathcal{T}), \\
        \mathcal{F}_t &= CrossAttn(e_t, \mathcal{F}_{hidden}), \\
        e_i &= Connector(\text{PGFT-CLIPV}(\mathcal{I})), \\
        \mathcal{F}_i &= CrossAttn(e_i, \mathcal{F}_{hidden}), \\
    \end{aligned}
\end{equation}
where $\mathcal{T}$ and $\mathcal{I}$ are the input text and image, $e_t$ and $e_i$ are the text and image embeddings, CLIPT is the CLIP text encoder, $\mathcal{F}_t$ and $\mathcal{F}_i$ are the outputs of the cross-attention mechanism $CrossAttn$, $Connector$ is the linear connector and $\mathcal{F}_{hidden}$ are the latent representations of the T2I model.

Guided by prominent annotation models, the textual branch primarily encodes high-level representations, whereas the visual branch complements it with low-level cues.
To adaptively balance their contributions, LGWAM is introduced to modulate the influence of the visual branch within DPCA.
It comprises lightweight projection layers that scale the output of the cross-attention mechanism.
The scaled low-level visual representations are fused with the high-level textual representations from the other branch.
We define the process as follows.
\begin{equation}
    \begin{aligned}
        \mathcal{F}_i' &= G(\mathcal{F}_i), \\
        \mathcal{F}' &= MLP(\mathcal{F}_i' + \mathcal{F}_t) + \mathcal{F}_{hidden}, \\
    \end{aligned}
\end{equation}
where $\mathcal{F}_i'$ are the scaled representations, $G$ is the LGWAM, $\mathcal{F}'$ is the fused cross-level semantic representations and $MLP$ is the multi-layer perceptron.

With DPCA, the proposed method is enriched with fine-grained visual semantics.
With LGWAM, high-level textual semantics and low-level visual attributes are dynamically weighted across blocks of different widths.
Finally, the CSM facilitates reconstruction with coherent dual-branch integration and enhanced semantic richness.

\subsection{Multimodal Signal Fusion Module}
\label{subsec:module2}

Given the information dimensionality and characteristics of different signals, we introduce a Multimodal Signal Fusion Module (MSFM) to progressively integrate three signals into the T2I models, which is shown in \autoref{fig:framework}.

In the first stage, since depth maps and semantic segmentation masks provide complementary vertical and horizontal structural cues along the channel dimension, we concatenate them first and extract shallow representations of the concatenated result with a convolutional block.
To enhance the interaction between depth maps and semantic segmentation masks, we pass the shallow representations through a channel attention mechanism, which then yields the multi-dimensional signals as:
\begin{equation}
    \begin{aligned}
        \mathcal{G}_c &= Concat(\mathcal{G}_d, \mathcal{G}_s), \\
        \mathcal{F}_c &= ChannelAttn(Conv(\mathcal{G}_c)), \\
        \mathcal{G}_r &= \mathcal{G}_c \otimes \mathcal{F}_c, \\
    \end{aligned}
    \label{equ:MS-GAM}
\end{equation}
where $\mathcal{G}_d$, $\mathcal{G}_s$, $\mathcal{G}_c$, and $\mathcal{G}_r$ are depth maps, semantic segmentation masks, concatenated signals, and multi-dimensional signals, respectively.
$\mathcal{F}_c$ are shallow representations, $Concat$ is the concatenation operation, $ChannelAttn$ is the channel attention mechanism, $Conv$ is the convolutional block, and $\otimes$ is the element-wise multiplication.

In the second stage, we modulate the HED boundaries and multi-dimensional signals to the T2I models in a dual-path manner.
Specifically, both the HED boundaries and the multi-dimensional signals are processed by separate semantic adaptive feature transformation blocks (SAFT)~\cite{park_saft_2019} to scale and shift the representations of the T2I model.
Then, they are fused via element-wise addition, allowing information integration while preserving the individual signal contributions.
The process is formulated as:
\begin{equation}
    \begin{aligned}
        \mathcal{F}_h &= SAFT(Norm(\mathcal{F}_{hidden}), \mathcal{G}_h), \\
        \mathcal{F}_r &= SAFT(Norm(\mathcal{F}_{hidden}), \mathcal{G}_r), \\
        \mathcal{F}' &= \mathcal{F}_h \oplus \mathcal{F}_r, \\
    \end{aligned}
    \label{equ:MS-GAM2}
\end{equation}
where $\mathcal{G}_h$ are the HED boundaries, $\mathcal{F}_h$ and $\mathcal{F}_r$ are the transformed outputs of SAFT blocks, $Norm$ is the Layer Normalization, $\mathcal{F}'$ are the fused latent representations, and $\oplus$ is the element-wise addition.

By integrating HED boundaries, depth maps, and semantic segmentation masks within the proposed MSFM, the T2I models are enriched with multi-dimensional dense guidance, improving the consistency of the outcomes.

%% file: parts/V_experiments.tex
\section{Experiments}
\label{sec:exp}

\input{tables/implementation_details}

\input{tables/main_table_II}

\subsection{Experimental Settings}
\label{subsec:exp_settings}

\subsubsection{Datasets and Implementation Details}
\label{subsubsec:datasets}

\textit{Real-ISR tasks.}
We trained the proposed model on general image datasets, including LSDIR~\cite{li_lsdir_2023} and the first 10,000 images from FFHQ~\cite{bai_ffhq_2023}, combining with the degradation pipeline proposed by Real-ESRGAN~\cite{wang_realesrgan_2021} to generate LR-HR image pairs.
For evaluation, we validated the performance of the method on six benchmarks, which include real-world datasets like DPED-iPhone~\cite{ignatov_dpediphone_2017}, RealLR200~\cite{wu_seesr_2024}, and RealLQ250~\cite{ai_reallq250_2024}, as well as synthetic datasets like RealSR~\cite{cai_realsr_2019}, DIV2K-Val~\cite{agustsson_div2kval_2017} and DRealSR~\cite{wei_drealsr_2020}.

We conducted the experiment on 2 NVIDIA A100 40G GPUs, with 512 $\times$ 512 resolution HR images and 128 $\times$ 128 resolution LR images.
Detailed experimental settings are shown in \autoref{tab:implementation_details}.





\noindent \textit{Prior-guided fine-tuning.}
We fine-tuned the signal extractors with datasets from their original setups or general image collections.
Specifically, for the HED and segmentation modalities, we selected LSDIR~\cite{li_lsdir_2023}, DF2K~\cite{lim_df2k_2017}, and FFHQ~\cite{bai_ffhq_2023}.
For depth modality, we incorporated LSDIR~\cite{li_lsdir_2023}, DF2K~\cite{lim_df2k_2017} and NYU\_Depth\_V2~\cite{Silberman_nyudepthv2_2012}.
And for the CLIP vision encoder, we employed ImageNet-1K~\cite{russakovsky_imagenet1k_2015}.

Following the method described in \autoref{subsec:module0}, we fine-tuned HED~\cite{xie_hed_2017} for edge detection, MaskDINO~\cite{li_maskdino_2023} for semantic segmentation, DepthAnythingV2~\cite{yang_depthanythingv2_2024} for depth estimation, and CLIP ViT-H/14 \cite{radford_clip_2021} for image embedding extraction.
All experiments were conducted on an NVIDIA A100 40G GPU.
And detailed experimental settings are shown in \autoref{tab:implementation_details}.

\subsubsection{Evaluation Metrics}
\label{subsubsec:eva_metrics}

To enable a comprehensive comparison with contemporary methods, we adopted seven widely used evaluation metrics, including both reference-based and non-reference ones.
In general, reference-based metrics require a ground-truth image for comparison, while non-reference metrics evaluate directly from the generated image.
For image fidelity, we utilized reference-based metrics, including PSNR and SSIM \cite{wang_ssim_2004}, with the calculations conducted in the YCbCr space.
For perceptual quality, we utilized the remaining five metrics, including LPIPS \cite{zhang_lpips_2018} as reference-based metrics, and NIQE \cite{zhang_niqe_2015}, MANIQA \cite{yang_maniqa_2022}, MUSIQ \cite{ke_musiq_2021}, and CLIPIQA \cite{wang_clipiqa_2023} as non-reference metrics.
Notably, all evaluations were conducted with the IQA-Pytorch project~\cite{chen_pyiqa_2022} to ensure fairness.


\input{tables/main_table_I}

\subsection{Comparison to State-of-the-Art Methods}
\label{subsec:cmp}

\subsubsection{Quantitative Comparison}
\label{subsubsec:quant_cmp}

We presented a comprehensive quantitative comparison between the proposed method and several state-of-the-art methods.
As shown in \autoref{tab:main_table_II}, for real-world benchmarks, the proposed method excels in human perception-based evaluation metrics.
It achieves the best MANIQA~\cite{yang_maniqa_2022} performance in all datasets, surpassing the second-best approaches by 1.6\% on DPED-iPhone~\cite{ignatov_dpediphone_2017}, 3.9\% on RealLR200~\cite{wu_seesr_2024} and 0.8\% on RealLQ250~\cite{ai_reallq250_2024}, respectively.
Moreover, MegaSR exhibits a 2.0\% improvement in CLIPIQA~\cite{wang_clipiqa_2023} scores on DPED-iPhone~\cite{ignatov_dpediphone_2017}, while maintaining competitive results on other metrics.
The results highlight the superiority of MegaSR in real-world scenarios.

As shown in \autoref{tab:main_table_I}, for synthetic benchmarks, the proposed method maintains its advantage on quality-driven metrics, where GAN-based methods perform poorly.
Notably, MegaSR achieves the best MANIQA~\cite{yang_maniqa_2022} and CLIPIQA~\cite{wang_clipiqa_2023} scores on RealSR~\cite{cai_realsr_2019}, surpassing the second-best method by 2.5\% and 1.3\%, respectively.
Additionally, MegaSR gains a robust improvement on the larger scale DIV2K-Val~\cite{agustsson_div2kval_2017}, achieving the best MUSIQ~\cite{ke_musiq_2021} score while maintaining MANIQA~\cite{yang_maniqa_2022} and CLIPIQA~\cite{wang_clipiqa_2023} at a competitive second place.
Besides quality-driven metrics, MegaSR also exhibits competitive performance among T2I-based methods in terms of fidelity-focused metrics, where GAN-based counterparts outperform.
This trend is consistent with other benchmarks, which demonstrates the advantage of enhancing perceptual quality while maintaining fidelity.

The above observations reinforce a well-established trade-off between fidelity and realism in previous studies~\cite{wu_seesr_2024, tsao_holisdip_2024}.
Although fidelity-focused metrics measure pixel differences, they tend to favor smoother results.
However, T2I-based methods generate visually plausible details that may not align with HRs.
This leads to discrepancies in performance when measured by these two types of metrics.

\subsubsection{Qualitative Comparison}
\label{subsubsec:qual_cmp}
\begin{figure*}[t]
    \centering
    \includegraphics[width=\linewidth]{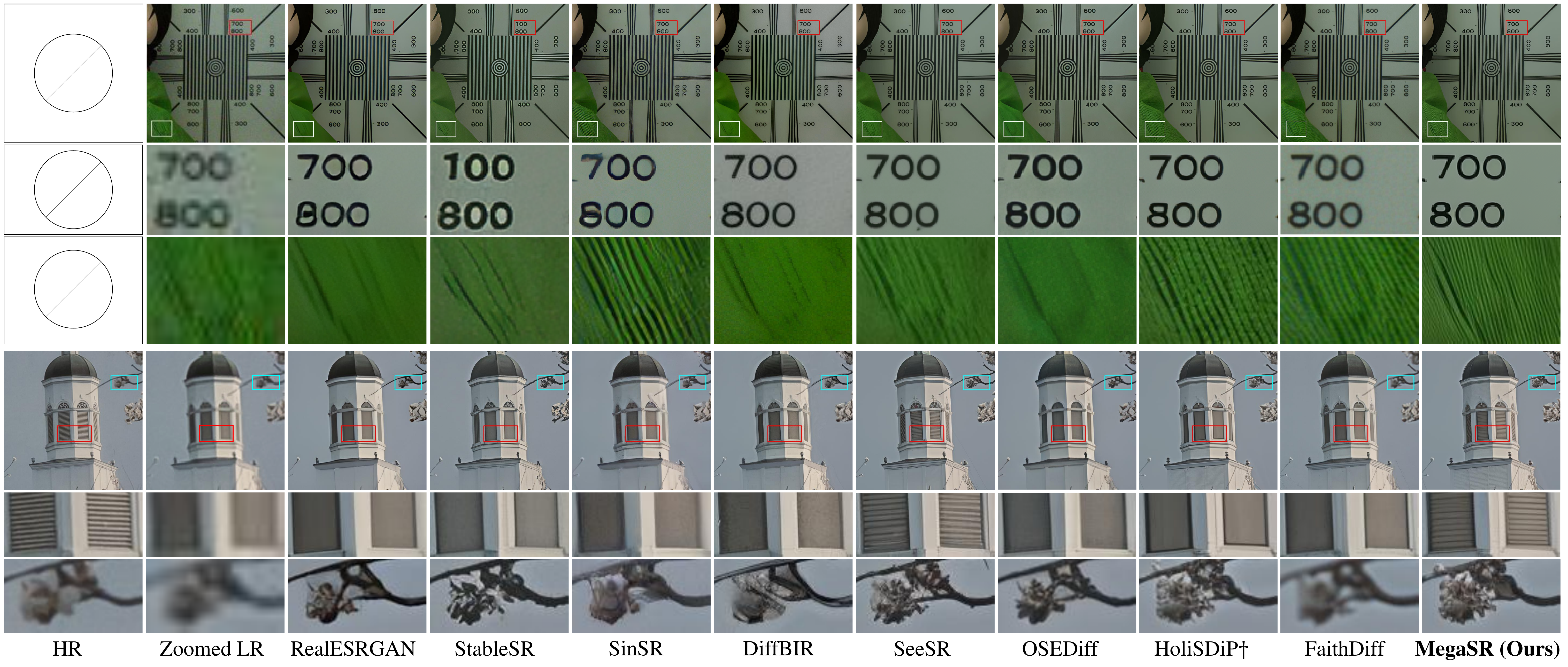}
    \vspace{-2em}
    \caption{Qualitative comparisons with different Real-ISR methods.
    The proposed method achieves superior fidelity and realism in terms of semantic preservation and structural consistency.}
    \label{fig:qualitative}
    \vspace{-1em}
\end{figure*}

\autoref{fig:qualitative} provides a qualitative visual comparison between the proposed method and the existing Real-ISR methods.
Specifically, in real-world scenarios, our method outperforms other GAN-based and T2I-based methods in terms of structural consistency.
For the text reconstruction results shown in the second row, RealESRGAN~\cite{wang_realesrgan_2021}, StableSR~\cite{wang_stablesr_2024}, SinSR~\cite{wang_sinsr_2024}, DiffBIR~\cite{lin_diffbir_2024}, OSEDiff~\cite{wu_osediff_2024}, and HoliSDiP~\cite{tsao_holisdip_2024} introduce structural distortions of the numbers, while SeeSR~\cite{wu_seesr_2024} and FaithDiff~\cite{chen_faithdiff_2025} struggle with clarity.
When it comes to texture generation shown in the third row, RealESRGAN, StableSR, DiffBIR, SeeSR, and OSEDiff tend to produce overly smoothed results, whereas SinSR and HoliSDiP exhibit unnatural patterns.

In synthetic scenarios, the proposed method maintains its superiority.
As shown in the fifth row, Real-ESRGAN generates visually smooth results for windows, while introducing unpleasant artifacts in the reconstruction of flowers.
Although other T2I-based methods achieve greater realism, with the exception of SeeSR, they also struggle to reconstruct fine-grained details for windows.
While SeeSR captures these details, it lacks structural neatness.
For semantic consistency, as shown in the sixth row, the proposed method demonstrates an advantage in semantic preservation and clarity.
RealESRGAN, SinSR, and DiffBIR fail to preserve the semantics of the flowers, resulting in unrecognizable objects.
Although StableSR, SeeSR, OSEDiff, and HoliSDiP mitigate it to some extent, they exhibit noticeable artifacts.
FaithDiff also performs poorly in preserving the clarity of fine objects.

\input{tables/ablation_ft}
\input{tables/ablation_extractors}
\input{tables/ablation_extractor_parameters}
\input{tables/ablation_dpca}

\subsection{Ablation Studies}
\label{subsec:ablation}

In this section, we evaluated the effectiveness and sensitivity of different components in our proposed method.
We first assessed the impact of prior-guided fine-tuning strategies and analyzed the sensitivity of extractors in \autoref{subsubsec:ablation_pgft}.
We then tested the contribution of each component within the CSM in \autoref{subsubsec:ablation_csm}.
Next, we analyzed the multimodal signals and their sensitivity in \autoref{subsubsec:ablation_msfm}.

\subsubsection{Effectiveness of Prior-Guided Fine-Tuning Strategies}
\label{subsubsec:ablation_pgft}

(1) \textit{Effectiveness of the prior-guided fine-tuning:}
In \autoref{subsec:module0}, we introduced different fine-tuning strategies to extractors with different degradation priors.
We further investigated their impact by conducting three ablation experiments:
a) directly utilizing the pre-trained extractors (Ours w/o ft);
b) full-parameter fine-tuning the depth modality~\cite{yang_depthanythingv2_2024} (Ours w/ depth-full); and
c) full-parameter fine-tuning the CLIP vision encoder~\cite{radford_clip_2021} (Ours w/ CLIPV-full).
The results are shown in \autoref{tab:ablation_ft}.
First, when directly utilizing the pre-trained extractors, a noticeable decrease is observed in both fidelity-focused and quality-driven metrics.
The fine-tuning enhances their ability to generate accurate signals towards LR inputs.
Secondly, while full-parameter fine-tuning the depth modality and the CLIP vision encoder partially disrupts the degradation priors, the impact remains marginal.

\begin{figure*}[ht]
    \centering
    \begin{minipage}[t]{0.48\linewidth}
        \centering
        \includegraphics[width=\linewidth]{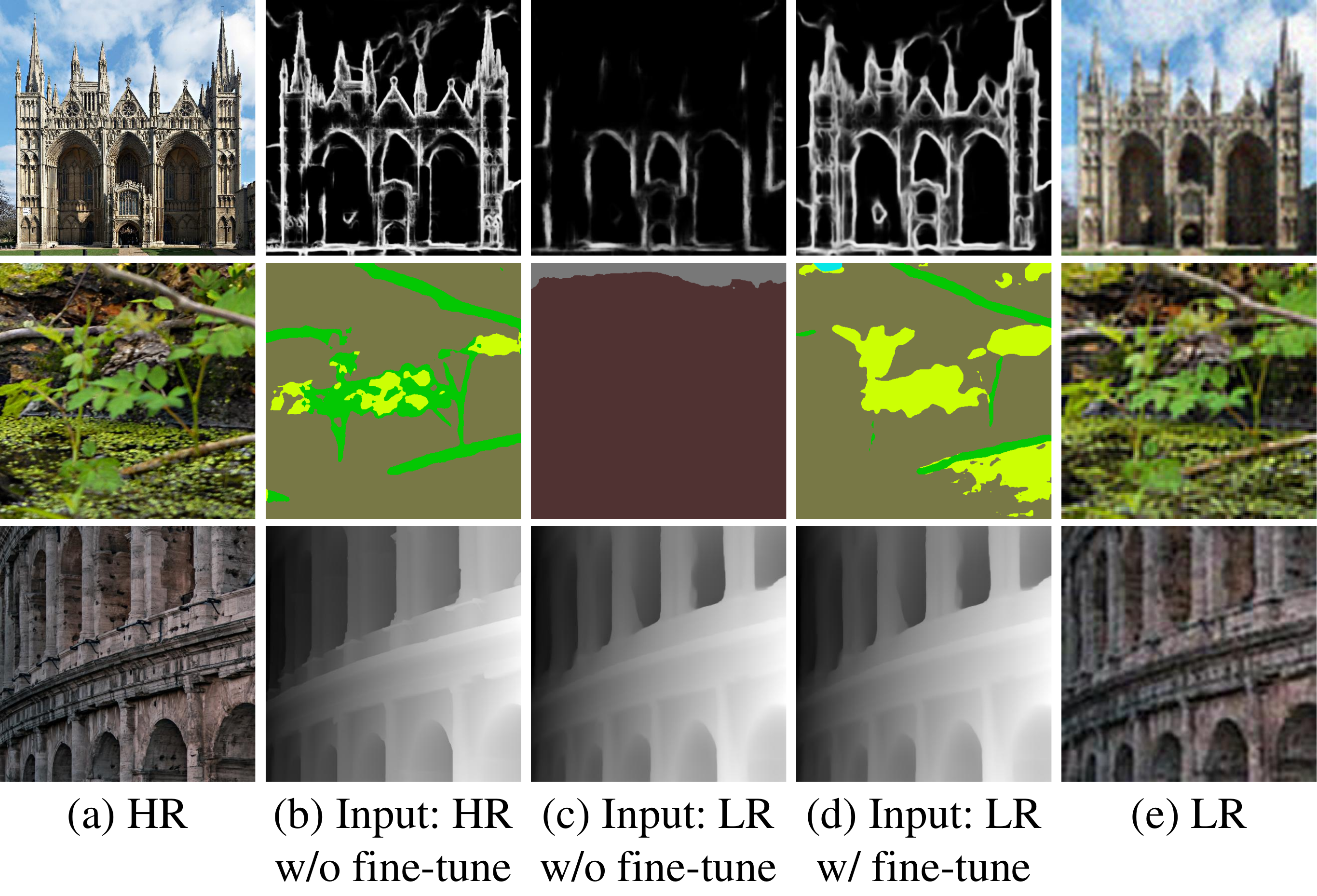}
        \vspace{-2em}
        \caption{Visual comparison of high-level outputs before and after fine-tuning with LR and HR images as input, respectively.}
        \vspace{-1em}
        \label{fig:ablation_ft}
    \end{minipage}
    \hfill
    \begin{minipage}[t]{0.48\linewidth}
        \centering
        \includegraphics[width=\linewidth]{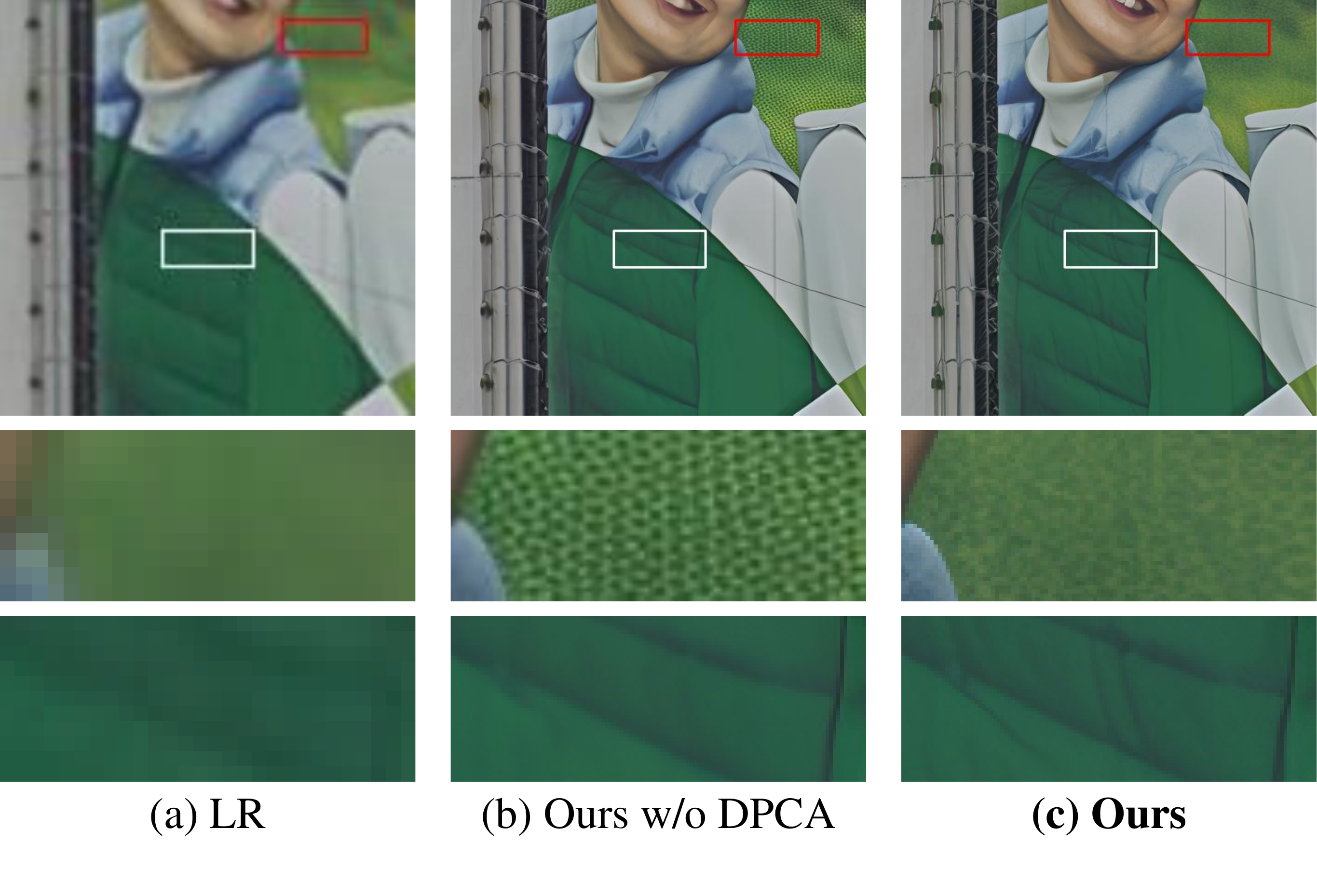}
        \vspace{-2em}
        \caption{Visualization of the effectiveness of DPCA. DPCA enhances the representations and facilitates high-fidelity reconstruction.}
        \vspace{-1em}
        \label{fig:ablation_dpca}
    \end{minipage}
\end{figure*}

\input{tables/ablation_lgwam}

To visualize the effectiveness of PGFT, we fed both LR and HR images into the models before and after fine-tuning and then compared the results.
As shown in \autoref{fig:ablation_ft}, the improvements are obvious the for HED~\cite{xie_hed_2017} and segmentation~\cite{li_maskdino_2023} modalities.
However, the depth~\cite{yang_depthanythingv2_2024} modality already performs well initially, and fine-tuning further enhances the accuracy of fine details.
For the CLIP vision encoder~\cite{radford_clip_2021}, we computed the cosine similarity of the LR embeddings to the HR counterparts before and after fine-tuning on the RealSR dataset.
The similarity improves slightly from 0.8217 to 0.8401, indicating a moderate enhancement that aligns with the trend observed in the depth modality.
In summary, the HED and segmentation modalities exhibit fewer degradation priors, making them more suitable for full-parameter fine-tuning to fully exploit their capacity.
In contrast, the depth modality and CLIP vision encoder retain stronger priors, and thus LoRA fine-tuning is applied to enhance fine details while maintaining efficiency.

(2) \textit{Extractor Sensitivity Analysis:}
We replaced the signal extractors with different large-scale visual models to validate the flexibility of the proposed method and framework.
Specifically, we replaced DepthAnythingV2~\cite{yang_depthanythingv2_2024} with ZoeDepth \cite{bhat_zoedepth_2023}, MaskDINO~\cite{li_maskdino_2023} with MaskFormer \cite{cheng_mask2former_2022} and the CLIP vision encoder~\cite{radford_clip_2021} with Swin Transformer V2 \cite{liu_swintransformerv2_2022}, respectively.
The results are presented in \autoref{tab:ablation_extractors}.
We found that replacing the extractors results in only minor differences compared to the original setup.
This indicates that the performance gains arise primarily from the multimodal guidance signals rather than the extractors themselves.
Any extractor capable of capturing such signals can readily adapt to our method, leaving room for further exploration.

We further discussed the overall parameters and computational overhead of different extractors in \autoref{tab:ablation_extractor_parameters}.
Notably, although these extractors introduce additional parameters and computational costs, the increases remain within an acceptable range.
Based on these observations, our implementation employs the state-of-the-art extractors while maintaining computational efficiency.

\subsubsection{Effectiveness of CSM}
\label{subsubsec:ablation_csm}

(1) \textit{Effectiveness of the DPCA.}
The proposed DPCA extracts fine-grained visual semantic knowledge and injects it into the T2I model to mitigate the fine detail deficiency.
We removed the visual branch within DPCA and compared the performance with the original settings.
As shown in \autoref{tab:ablation_dpca}, our full method outperforms the variant without DPCA in quality-driven metrics, notably improving MANIQA \cite{yang_maniqa_2022} from 0.5371 to 0.5574, MUSIQ \cite{ke_musiq_2021} from 69.3150 to 70.0252, and CLIPIQA \cite{wang_clipiqa_2023} from 0.6553 to 0.6790 on the RealSR dataset, while remaining competitive in other metrics.

Despite slightly lower PSNR values, the proposed method yields visually superior results.
As shown in \autoref{fig:ablation_dpca}, the full model avoids the over-smoothing and fine-detail artifacts observed in the variant output.
It demonstrates that DPCA enhances image-level contextual representations and facilitates high-fidelity reconstruction.

\begin{figure}[t]
    \centering
    \includegraphics[width=\linewidth]{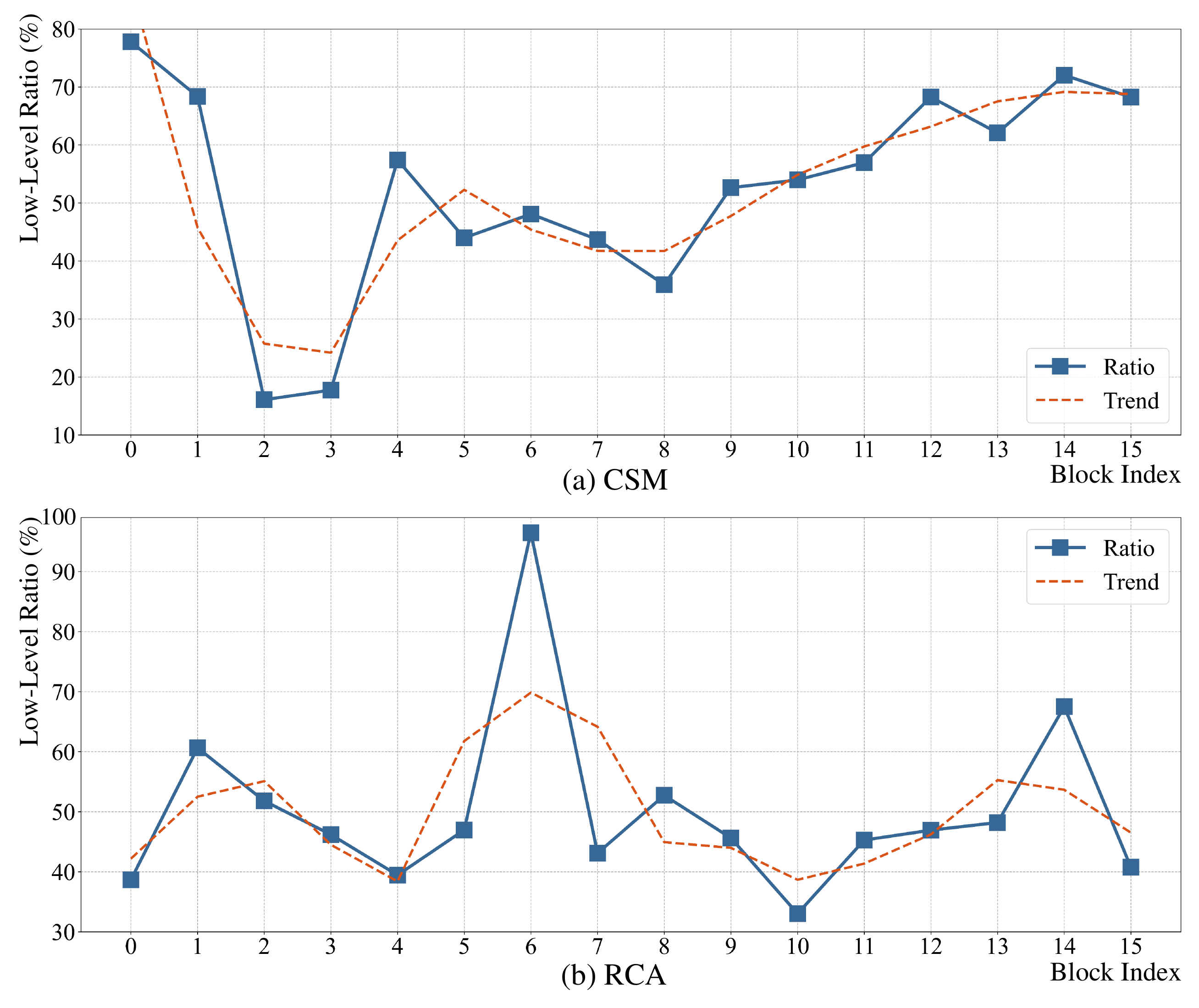}
    \vspace{-2em}
    \caption{Visualization of the ratio between the weights of low-level attributes and high-level semantics across each block in the CSM and the RCA of SeeSR on the RealSR dataset.}
    \vspace{-1.7em}
    \label{fig:ablation_ratios}
\end{figure}

\begin{figure*}[t]
    \centering
    \includegraphics[width=\linewidth]{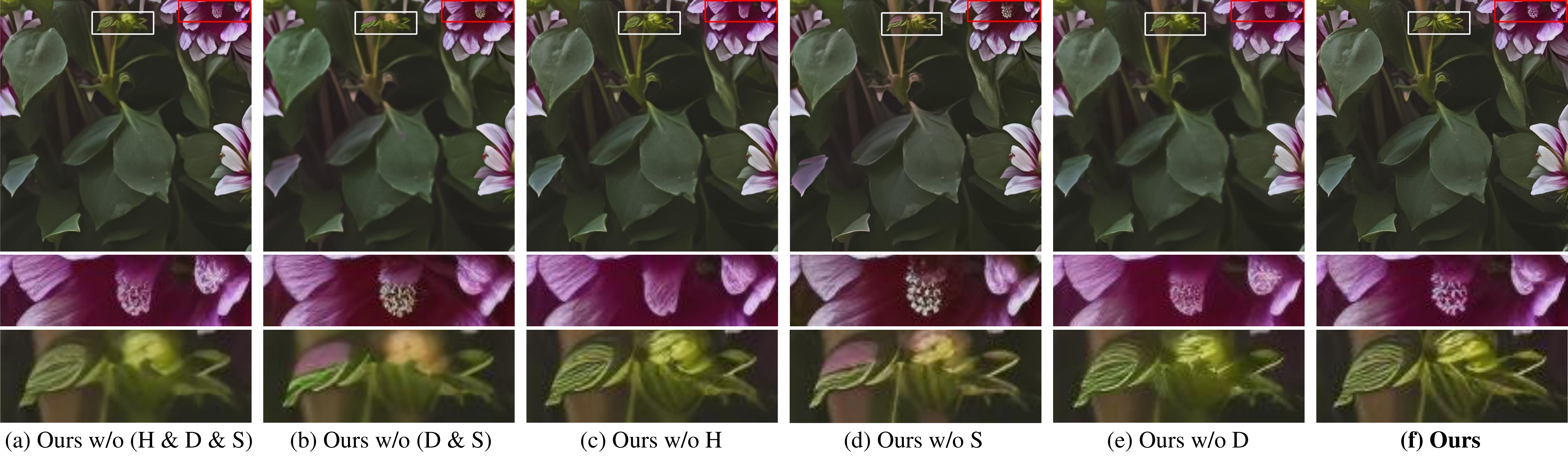}
    \vspace{-1em}
    \caption{
    Visualization of different combinations of guidance signals and their effects on the results.
    The HED boundaries improve object delineation, depth maps enhance color consistency, and segmentation masks refine texture details.
    }
    \label{fig:ablation_signals}
    \vspace{-1em}
\end{figure*}

(2) \textit{Effectiveness of the LGWAM.}
The LGWAM adaptively scales the visual semantic representations in DPCA, ensuring that blocks across different widths within the T2I U-Net receive appropriate semantic knowledge.
We ablated LGWAM in DPCA and present the results in \autoref{tab:ablation_lgwam}.
Comparing the second and fourth rows, removing the LGWAM leads to a noticeable drop in perceptual quality, whereas the improvement in fidelity is minor.
This is because, without the modulation, the model struggles to effectively regulate the weights of different semantic controls.
As a result, for example, narrow blocks also receive a large amount of low-level knowledge, leading to degraded performance.

\input{tables/ablation_signals}

To further analyze the modulation behavior, we calculated the ratio between the weights of the low-level attributes and the high-level semantics across each block in both our proposed CSM and the RCA module of SeeSR~\cite{wu_seesr_2024} on the RealSR dataset.
A higher ratio indicates a greater emphasis on image-level attributes.
The results are shown in \autoref{fig:ablation_ratios}.
Firstly, the low-level ratio of the CSM exhibits a high–low–high distribution pattern across U-Net blocks, suggesting that the blocks at both ends focus more on fine-grained low-level attributes, whereas the middle blocks increasingly attend to high-level semantic knowledge.
This observation is consistent with our analysis of semantic requirements in different blocks.
In contrast, the trend of the RCA appears more like a fitted pattern than a control of the proportions of different semantics across different blocks.

Building upon this observation, a natural question arises: would simply adding LGWAM to the RCA module suffice to achieve similar per-block semantic adaptation?
We conducted an experiment to verify this and present the results in \autoref{tab:ablation_lgwam}.
Comparing the third and fourth rows, there is a noticeable decrease in MANIQA~\cite{yang_maniqa_2022} and MUSIQ~\cite{ke_musiq_2021}.
Given the architectural differences between our method and SeeSR, the distinct designs for high- and low-level semantics may account for this performance drop.
In SeeSR, the TCA module is responsible for integrating textual semantics, while the RCA module fuses visual semantics.
Firstly, these two modules handle semantic control mutually.
The input to the RCA module already contains representations in which the semantic control from TCA has been embedded.
Secondly, TCA and RCA operate as two serial switches, which inherently complicates the regulation of semantic control.
Although LGWAM can provide stronger modulation, it still struggles to effectively control the overall generative behavior.
In contrast, our implementation adopts a dual-branch fusion mechanism.
On the one hand, it simplifies the control process as the two branches operate without interfering with each other.
On the other hand, it reduces the disturbance to the generative priors of the T2I model.

\input{tables/ablation_canny_sketch}

\subsubsection{Effectiveness of MSFM}
\label{subsubsec:ablation_msfm}

\begin{figure*}[t]
    \centering
    \includegraphics[width=\linewidth]{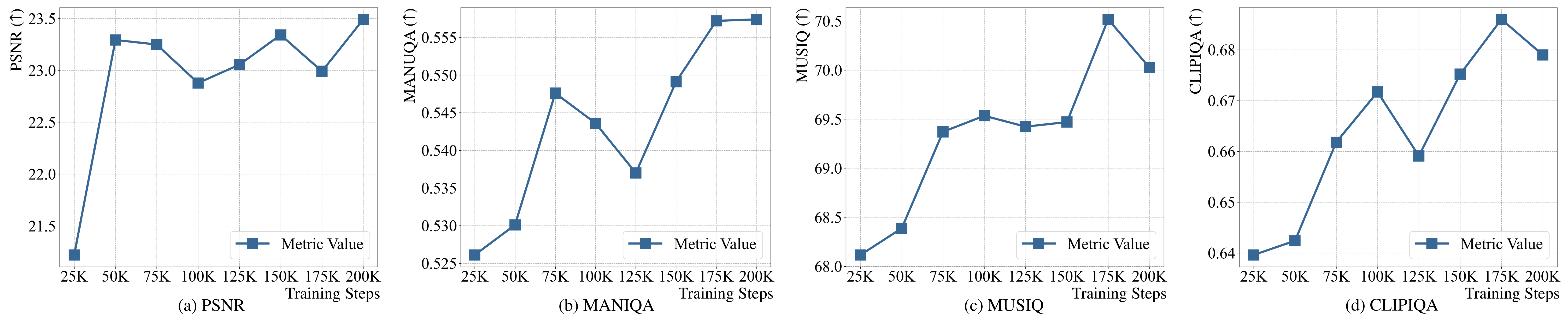}
    \vspace{-1.5em}
    \caption{
    Visualization of how different metrics evolve over training steps.
    The metrics generally improve as training progresses.
    Given both image fidelity and perceptual realism, we selected the results at 200K steps as the final prediction.
    }
    \label{fig:training_process}
    \vspace{-1em}
\end{figure*}

(1) \textit{Effectiveness of the signals:}
MSFM is proposed to incorporate guidance signals into the diffusion process.
To evaluate the contribution of different guidance signals, we conducted comparative experiments that selectively excluded each component.
As shown in \autoref{tab:ablation_signals}, incorporating HED boundaries~\cite{xie_hed_2017} enhances realism and maintains fidelity, as reflected by the gains in MANIQA~\cite{yang_maniqa_2022} and MUSIQ~\cite{ke_musiq_2021}, with only a minor decrease in PSNR.
Comparing the second, fourth, and fifth rows, the depth maps~\cite{yang_depthanythingv2_2024} and segmentation masks~\cite{li_maskdino_2023} further enhance semantic alignment with textual tags, achieving an improvement in CLIPIQA~\cite{wang_clipiqa_2023}.
Our full model achieves the best trade-off between fidelity and realism, enabling richer details through a slight compromise in fidelity.
An interesting observation is that adding either depth or segmentation modality individually leads to performance drops, while combining both gives strong gains.
We attribute this phenomenon to the different qualities of the signals involved.
As shown in \autoref{fig:ablation_ft}, due to the limitations of existing methods, the quality of depth maps and segmentation masks is not as good as the HED boundaries even after fine-tuning.
As a result, using only depth maps or segmentation masks can lead to performance degradation.
However, these two modalities contain unique detailed information, which is crucial for enhancing fine textures in Real-ISR and cannot be replaced by other modalities.
To mitigate the impact of these issues while preserving their distinctive information, we designed the MSFM to employ these two modalities to supplement the HED modality, thereby achieving the final enhancement.

\autoref{fig:ablation_signals} provides qualitative evidence to support these findings.
First, a comparison between (b), (c), and (f) indicates that HED boundaries enhance the object delineation of fine-grained details.
In contrast, (d), (e), and (f) demonstrate that while depth maps primarily improve the color consistency, segmentation masks play a complementary role by enhancing the texture details of semantic objects.
The complete integration of the signals facilitates the model to capture comprehensive visual attributes, thus improving texture realism and structural consistency.

(2) \textit{Signal Sensitivity Analysis:}
In \autoref{subsec:signals}, we selected HED boundaries~\cite{xie_hed_2017}, depth maps~\cite{yang_depthanythingv2_2024}, and semantic segmentation masks~\cite{li_maskdino_2023} to exert dense guidance on the proposed method.
To further investigate the impact of other signals on performance gains, we conducted experiments by adding Canny edges and sketch maps.

The results are presented in \autoref{tab:ablation_canny_sketch}.
The integration of these two signals leads to marginal improvements.
This is because, for each category, only the one with the strongest intensity was selected so that the information would be rich and minimally redundant.
The HED boundaries, Canny edges, and sketch maps all belong to edge-based signals, while the HED boundaries exhibit the strongest intensity among them.
Adding other redundant guidance signals would increase the computational cost without providing a corresponding improvement in performance.

\input{tables/overall_parameters}

\subsection{Model Efficiency}
\label{subsec:overall_analyses}
\subsubsection{Training Curve Analysis}
\label{subsubsec:training_curve}
We visualized how different metrics evolve with the number of training steps.
As shown in \autoref{fig:training_process}, the metrics generally improve as training progresses.
Specifically, PSNR increases significantly during the early stages, then exhibits oscillations, and reaches its peak at 200K steps.
In contrast, MANIQA~\cite{yang_maniqa_2022} shows an overall upward trend, indicating steady improvements in perceptual quality.
The trend is consistent with MUSIQ~\cite{ke_musiq_2021} and CLIPIQA~\cite{wang_clipiqa_2023}, though a slight decline is observed in the final stage.
Given both image fidelity and perceptual realism, we selected the results at 200K steps as the final prediction.

\subsubsection{Model Parameters}
\label{subsubsec:model_parameters}
\autoref{tab:overall_parameters} compares the total parameters, trainable parameters, and runtime of different Real-ISR methods in generating $512 \times 512$ images from $128 \times 128$ inputs.
To ensure a fair comparison, we calculated the average time on the DIV2K-Val dataset~\cite{agustsson_div2kval_2017}, which contains 3,000 images in total.
The batch size was set to 1.
And all experiments were conducted on an NVIDIA A100 40G GPU.

First, GAN-based methods are generally more efficient compared to T2I-based methods, primarily due to their smaller model sizes and the ability to generate images in a single forward pass.
However, they tend to produce unrecognizable objects and introduce unpleasant artifacts.
Second, among T2I-based methods that share the same base model, SeeSR~\cite{wu_seesr_2024} has the fewest parameters, followed by HoliSDiP~\cite{tsao_holisdip_2024} and the proposed method.
Since StableSR~\cite{wang_stablesr_2024} and DiffBIR~\cite{lin_diffbir_2024} utilize different base models, we exclude them from the parameter comparison.
Third, StableSR achieves the fastest runtime among the T2I-based methods, demonstrating its efficiency.
Finally, although MegaSR achieves superior results, it has more parameters and a longer inference time compared to StableSR, SeeSR, and HoliSDiP.
We consider it a reasonable and necessary trade-off as the proposed method incorporates more signals to enhance the reconstruction quality.

%% file: tables/implementation_details.tex
\begin{table}[t]
    \renewcommand\arraystretch{1}
    \caption{
    Detailed experimental settings of the Real-ISR tasks and prior-guided fine-tuning.
    }
    \label{tab:implementation_details}
    \centering
    \resizebox{\linewidth}{!}{
        \begin{tabular}{ccccc}
            \Xhline{1.25pt}
            Models & Optim. & Batch Size & Learning Rate & Steps \\
            \Xhline{0.75pt}
            MegaSR & Adam & 16 & 2.5$e^{-5}$ & 200K \\
            HED~\cite{xie_hed_2017} & Adam & 32 & 1$e^{-4}$ & 300K \\
            DepthAnythingV2~\cite{yang_depthanythingv2_2024} & Adam & 32 & 1$e^{-4}$ & 60K \\
            MaskDINO~\cite{li_maskdino_2023} & Adam & 32 & 1$e^{-4}$ & 85K \\
            CLIPV~\cite{radford_clip_2021} & Adam & 128 & $5e^{-5}$ & 25K \\
            \Xhline{1.25pt}
        \end{tabular}
    }
    \vspace{-1em}
\end{table}

%% file: tables/main_table_II.tex
\begin{table*}[htbp]
    \renewcommand\arraystretch{1}
    \newcommand{\first}[1]{\textcolor{red}{\textbf{#1}}}
    \newcommand{\second}[1]{\textcolor{blue}{\textbf{#1}}}
    \newcommand{\third}[1]{\underline{\textbf{#1}}}
    \centering
    \caption{Quantitative comparison with existing Real-ISR methods on real-world datasets.
    The best, second-best, and third-best results are highlighted in \first{red}, \second{blue}, and \third{black}, respectively.
    $\dagger$ indicates our reproduced version due to certain flaws in the original code.}
    \label{tab:main_table_II}
    \begin{tabular}{ccp{1.7cm}<{\centering}p{1.7cm}<{\centering}p{1.7cm}<{\centering}p{1.7cm}<{\centering}}
        \Xhline{1.25pt}
        Datasets & Method & NIQE$\downarrow$ & MANIQA$\uparrow$ & MUSIQ$\uparrow$ & CLIPIQA$\uparrow$ \\
        \Xhline{0.75pt}
        \multirow{13}{*}{\textit{DPED-iPhone}}
        & BSRGAN~\cite{zhang_bsrgan_2021} & \second{6.4908} & 0.3136 & 45.8906 & 0.4022 \\
        & Real-ESRGAN~\cite{wang_realesrgan_2021} & 6.8455 & 0.3036 & 42.4342 & 0.3382 \\
        & LDL~\cite{liang_ldl_2022} & 6.9554 & 0.3133 & 43.6455 & 0.3535 \\
        & FeMaSR~\cite{chen_femasr_2022} & \third{6.6287} & 0.3570 & 49.9494 & 0.5307 \\
        & DASR~\cite{liang_dasr_2022} & 6.6843 & 0.2557 & 32.6858 & 0.2826 \\
        & StableSR~\cite{wang_stablesr_2024} & 6.7322 & 0.3567 & 51.8923 & 0.4920 \\
        & SinSR~\cite{wang_sinsr_2024} & 7.9803 & 0.3569 & 46.7074 & 0.5743 \\
        & DiffBIR~\cite{lin_diffbir_2024} & 7.3028 & \third{0.4573} & 54.8280 & 0.5770 \\
        & SeeSR~\cite{wu_seesr_2024} & 6.7261 & \second{0.4581} & \first{57.6794} & \second{0.6077} \\
        & OSEDiff~\cite{wu_osediff_2024} & \first{6.3652} & 0.4425 & \second{56.3909} & \third{0.5927} \\
        & HoliSDiP$\dagger$~\cite{tsao_holisdip_2024} & 6.7430 & 0.4037 & 55.1995 & 0.5885 \\
        & FaithDiff~\cite{chen_faithdiff_2025} & 6.6519 & 0.3334 & 52.1251 & 0.4616 \\
        & \textbf{MegaSR} & 6.7751 & \first{0.4653} & \third{56.3767} & \first{0.6201} \\
        \Xhline{0.75pt}

        \multirow{13}{*}{\textit{RealLR200}}
        & BSRGAN~\cite{zhang_bsrgan_2021} & 4.3674 & 0.3704 & 64.8680 & 0.5700 \\
        & Real-ESRGAN~\cite{wang_realesrgan_2021} & 4.1771 & 0.3688 & 62.9601 & 0.5409 \\
        & LDL~\cite{liang_ldl_2022} & 4.3448 & 0.3730 & 63.1103 & 0.5364 \\
        & FeMaSR~\cite{chen_femasr_2022} & 4.6279 & 0.4077 & 64.2359 & 0.6548 \\
        & DASR~\cite{liang_dasr_2022} & 4.3181 & 0.2986 & 55.7096 & 0.4690 \\
        & StableSR~\cite{wang_stablesr_2024} & 4.2665 & 0.4126 & 67.5545 & 0.6848 \\
        & SinSR~\cite{wang_sinsr_2024} & 5.5981 & 0.4433 & 63.8286 & \first{0.7010} \\
        & DiffBIR~\cite{lin_diffbir_2024} & \first{3.9277} & 0.4647 & 66.7888 & \second{0.6979} \\
        & SeeSR~\cite{wu_seesr_2024} & \third{4.0822} & \third{0.4802} & 68.4247 & 0.6712 \\
        & OSEDiff~\cite{wu_osediff_2024} & \second{4.0162} & 0.4385 & \first{69.5941} & 0.6747 \\
        & HoliSDiP$\dagger$~\cite{tsao_holisdip_2024} & 4.2849 & \second{0.4842} & \third{68.9347} & 0.6727 \\
        & FaithDiff~\cite{chen_faithdiff_2025} & 4.3461 & 0.3594 & 63.1611 & 0.5803 \\
        & \textbf{MegaSR} & 4.2941 & \first{0.5032} & \second{69.1988} & \third{0.6892} \\
        \Xhline{0.75pt}

        \multirow{13}{*}{\textit{RealLQ250}}
        & BSRGAN~\cite{zhang_bsrgan_2021} & 4.5372 & 0.3514 & 63.5182 & 0.5690 \\
        & Real-ESRGAN~\cite{wang_realesrgan_2021} & \second{4.1293} & 0.3564 & 62.5145 & 0.5435 \\
        & LDL~\cite{liang_ldl_2022} & 4.2971 & 0.3598 & 62.2198 & 0.5446 \\
        & FeMaSR~\cite{chen_femasr_2022} & 4.2962 & 0.3358 & 61.8506 & 0.6216 \\
        & DASR~\cite{liang_dasr_2022} & 4.7857 & 0.2789 & 53.0230 & 0.4631 \\
        & StableSR~\cite{wang_stablesr_2024} & \third{4.1524} & 0.4042 & 67.3318 & 0.6859 \\
        & SinSR~\cite{wang_sinsr_2024} & 5.8143 & 0.4189 & 64.0219 & \first{0.7055} \\
        & DiffBIR~\cite{lin_diffbir_2024} & 5.5883 & 0.4115 & 59.9122 & 0.6272 \\
        & SeeSR~\cite{wu_seesr_2024} & 4.3301 & \second{0.4704} & \second{69.0847} & \third{0.6886} \\
        & OSEDiff~\cite{wu_osediff_2024} & \first{3.9671} & 0.4230 & \first{69.5524} & 0.6721 \\
        & HoliSDiP$\dagger$~\cite{tsao_holisdip_2024} & 4.6425 & \third{0.4486} & 65.8609 & 0.6504 \\
        & FaithDiff~\cite{chen_faithdiff_2025} & 4.7691 & 0.3388 & 63.8415 & 0.5856 \\
        & \textbf{MegaSR} & 4.4331 & \first{0.4741} & \third{68.0105} & \second{0.6940} \\
        \Xhline{1.25pt}
    \end{tabular}
    \vspace{-1em}
\end{table*}

%% file: tables/main_table_I.tex
\begin{table*}[t]
    \renewcommand\arraystretch{1}
    \newcommand{\first}[1]{\textcolor{red}{\textbf{#1}}}
    \newcommand{\second}[1]{\textcolor{blue}{\textbf{#1}}}
    \newcommand{\third}[1]{\underline{\textbf{#1}}}
    \centering
    \caption{Quantitative comparison with existing Real-ISR methods on synthetic datasets.
    The best, second-best, and third-best results are highlighted in \first{red}, \second{blue}, and \third{black}, respectively.
    $\dagger$ indicates our reproduced version due to certain flaws in the original code.}
    \label{tab:main_table_I}
    \begin{tabular}{ccp{1.3cm}<{\centering}p{1.3cm}<{\centering}p{1.3cm}<{\centering}p{1.3cm}<{\centering}p{1.5cm}<{\centering}p{1.5cm}<{\centering}p{1.5cm}<{\centering}}
        \Xhline{1.25pt}
        Datasets & Method & PSNR$\uparrow$ & SSIM$\uparrow$ & LPIPS$\downarrow$ & NIQE$\downarrow$ & MANIQA$\uparrow$ & MUSIQ$\uparrow$ & CLIPIQA$\uparrow$ \\
        \Xhline{0.75pt}
        \multirow{13}{*}{\textit{RealSR}} 
        & BSRGAN~\cite{zhang_bsrgan_2021} & \third{24.75} & \second{0.7401} & \second{0.2656} & 5.6348 & 0.3758 & 63.2861 & 0.5116 \\
        & Real-ESRGAN~\cite{wang_realesrgan_2021} & 24.15 & \third{0.7363} & \third{0.2710} & 5.8027 & 0.3737 & 60.3665 & 0.4491 \\
        & LDL~\cite{liang_ldl_2022} & 23.76 & 0.7300 & 0.2750 & 5.9916 & 0.3794 & 60.9249 & 0.4559 \\
        & FeMaSR~\cite{chen_femasr_2022} & 23.51 & 0.7088 & 0.2937 & 5.7673 & 0.3629 & 59.0588 & 0.5406 \\
        & DASR~\cite{liang_dasr_2022} & \first{25.40} & \first{0.7458} & 0.3134 & 6.5455 & 0.2470 & 41.2030 & 0.3202 \\
        & StableSR~\cite{wang_stablesr_2024} & 23.14 & 0.6796 & 0.3000 & 5.9755 & 0.4244 & 65.6786 & 0.6184 \\
        & SinSR~\cite{wang_sinsr_2024} & 24.48 & 0.7076 & 0.3186 & 6.2835 & 0.3997 & 60.5986 & 0.6280 \\
        & DiffBIR~\cite{lin_diffbir_2024} & 23.91 & 0.6220 & 0.3732 & 6.0788 & 0.5022 & 64.9021 & 0.6507 \\
        & SeeSR~\cite{wu_seesr_2024} & 23.60 & 0.6947 & 0.3007 & \third{5.4008} & \second{0.5437} & \second{69.8200} & \second{0.6700} \\
        & OSEDiff~\cite{wu_osediff_2024} & 23.59 & 0.7071 & 0.2920 & 5.6341 & 0.4711 & \third{69.0892} & \third{0.6693} \\
        & HoliSDiP$\dagger$~\cite{tsao_holisdip_2024} & 23.70 & 0.6994 & 0.2977 & \second{5.3884} & \third{0.5290} & 68.9540 & 0.6639 \\
        & FaithDiff~\cite{chen_faithdiff_2025} & \second{24.91} & 0.7129 & \first{0.2562} & 6.0413 & 0.3907 & 64.2661 & 0.5857 \\
        & \textbf{MegaSR} & 23.49 & 0.6903 & 0.3072 & \first{5.3795} & \first{0.5574} & \first{70.0251} & \first{0.6790} \\
        \Xhline{1pt}

        \multirow{13}{*}{\textit{DIV2K-Val}} 
        & BSRGAN~\cite{zhang_bsrgan_2021} & \first{22.79} & 0.5908 & 0.3351 & 4.7513 & 0.3532 & 61.1963 & 0.5247 \\
        & Real-ESRGAN~\cite{wang_realesrgan_2021} & \third{22.60} & \first{0.5986} & \third{0.3112} & \first{4.6787} & 0.3790 & 61.0566 & 0.5277 \\
        & LDL~\cite{liang_ldl_2022} & 22.18 & \second{0.5926} & 0.3256 & 4.8557 & 0.3727 & 60.0398 & 0.5180 \\
        & FeMaSR~\cite{chen_femasr_2022} & 21.32 & 0.5536 & 0.3126 & \third{4.7415} & 0.3443 & 60.8291 & 0.5998 \\
        & DASR~\cite{liang_dasr_2022} & \second{22.75} & \third{0.5924} & 0.3543 & 5.0273 & 0.3164 & 55.1963 & 0.5036 \\
        & StableSR~\cite{wang_stablesr_2024} & 21.62 & 0.5340 & 0.3125 & 4.7644 & 0.4198 & 65.7148 & 0.6770 \\
        & SinSR~\cite{wang_sinsr_2024} & 22.51 & 0.5675 & 0.3244 & 6.0022 & 0.4225 & 62.7473 & 0.6482 \\
        & DiffBIR~\cite{lin_diffbir_2024} & 21.83 & 0.5033 & 0.3763 & 5.1361 & \first{0.5231} & 66.3764 & \third{0.6840} \\
        & SeeSR~\cite{wu_seesr_2024} & 21.97 & 0.5673 & 0.3194 & 4.8095 & \third{0.5036} & \second{68.6699} & \first{0.6936} \\
        & OSEDiff~\cite{wu_osediff_2024} & 22.05 & 0.5735 & \second{0.2942} & \second{4.7081} & 0.4411 & \third{67.9716} & 0.6680 \\
        & HoliSDiP$\dagger$~\cite{tsao_holisdip_2024} & 22.21 & 0.5742 & 0.3248 & 4.9206 & 0.4848 & 67.5952 & 0.6635 \\
        & FaithDiff~\cite{chen_faithdiff_2025} & 22.49 & 0.5645 & \first{0.2640} & 5.1585 & 0.3659 & 65.1001 & 0.6049 \\
        & \textbf{MegaSR} & 22.05 & 0.5664 & 0.3157 & 4.8258 & \second{0.5074} & \first{68.7602} & \second{0.6892} \\
        \Xhline{1pt}

        \multirow{13}{*}{\textit{DRealSR}} 
        & BSRGAN~\cite{zhang_bsrgan_2021} & \third{26.39} & 0.7739 & \third{0.2858} & \third{6.5400} & 0.3404 & 57.1682 & 0.5099 \\
        & Real-ESRGAN~\cite{wang_realesrgan_2021} & 26.28 & \third{0.7767} & \second{0.2819} & 6.6929 & 0.3440 & 54.2724 & 0.4520 \\
        & LDL~\cite{liang_ldl_2022} & 25.97 & \second{0.7839} & \first{0.2792} & 7.1427 & 0.3429 & 53.9454 & 0.4477 \\
        & FeMaSR~\cite{chen_femasr_2022} & 24.85 & 0.7247 & 0.3157 & \first{5.9096} & 0.3165 & 53.7099 & 0.5638 \\
        & DASR~\cite{liang_dasr_2022} & \first{27.24} & \first{0.7995} & 0.3099 & 7.5857 & 0.2808 & 42.4116 & 0.3812 \\
        & StableSR~\cite{wang_stablesr_2024} & 25.73 & 0.7178 & 0.3262 & 6.6791 & 0.3855 & 59.8183 & 0.6241 \\
        & SinSR~\cite{wang_sinsr_2024} & 25.84 & 0.7151 & 0.3701 & 6.9613 & 0.3838 & 55.4730 & 0.6394 \\
        & DiffBIR~\cite{lin_diffbir_2024} & 24.99 & 0.6035 & 0.4662 & 6.6915 & \third{0.4935} & 60.4019 & 0.6483 \\
        & SeeSR~\cite{wu_seesr_2024} & 26.01 & 0.7330 & 0.3485 & 6.9710 & \first{0.5290} & \second{64.4534} & \third{0.6816} \\
        & OSEDiff~\cite{wu_osediff_2024} & 25.89 & 0.7546 & 0.2967 & \second{6.4197} & 0.4655 & \first{64.7006} & \first{0.6966} \\
        & HoliSDiP$\dagger$~\cite{tsao_holisdip_2024} & 25.48 & 0.7446 & 0.3167 & 6.6601 & 0.4781 & 63.1544 & 0.6450 \\
        & FaithDiff~\cite{chen_faithdiff_2025} & \second{26.50} & 0.7240 & 0.2989 & 6.8838 & 0.3666 & 59.1474 & 0.5892 \\
        & \textbf{MegaSR} & 25.74 & 0.7367 & 0.3258 & 6.6276 & \second{0.5098} & \third{64.1473} & \second{0.6853} \\
        \Xhline{1.25pt}
    \end{tabular}
    \vspace{-1em}
\end{table*}

%% file: tables/ablation_ft.tex
\begin{table}[t]
    \renewcommand\arraystretch{1.2}
    \caption{
    Effectiveness of prior-guided fine-tuning.
    }
    \label{tab:ablation_ft}
    \centering
    \resizebox{\linewidth}{!}{
        \begin{tabular}{ccccc}
            \Xhline{1.25pt}
            Methods & PSNR$\uparrow$ & MANIQA$\uparrow$ & MUSIQ$\uparrow$ & CLIPIQA$\uparrow$ \\
            \Xhline{0.75pt}
            Ours w/o ft & 23.2260 & 0.5483 & 68.9973 & 0.6751 \\
            Ours w/ depth-full & 23.4380 & 0.5417 & 69.4528 & 0.6703 \\
            Ours w/ CLIPV-full & 23.4719 & 0.5540 & 69.8267 & 0.6741 \\
            \Xhline{0.75pt}
            \textbf{Ours} & \textbf{23.4909} & \textbf{0.5574} & \textbf{70.0252} & \textbf{0.6790} \\
            \Xhline{1.25pt}
        \end{tabular}
    }
\end{table}

%% file: tables/ablation_extractors.tex
\begin{table}[t]
    \renewcommand\arraystretch{1.24}
    \caption{
    Extractor Sensitivity Analysis.
    }
    \label{tab:ablation_extractors}
    \centering
    \resizebox{\linewidth}{!}{
        \begin{tabular}{ccccc}
            \Xhline{1.25pt}
            Methods & PSNR$\uparrow$ & MANIQA$\uparrow$ & MUSIQ$\uparrow$ & CLIPIQA$\uparrow$ \\
            \Xhline{0.75pt}
            Ours w/ ZoeDepth & \textbf{23.5132} & 0.5533 & 69.1352 & 0.6801 \\
            Ours w/ MaskFormer & 23.4834 & 0.5494 & \textbf{70.0560} & 0.6710 \\
            Ours w/ Swin & 23.5012 & 0.5495 & 69.7815 & 0.6817 \\
            \Xhline{0.75pt}
            \textbf{Ours} & 23.4909 & \textbf{0.5574} & 70.0252 & \textbf{0.6790} \\
            \Xhline{1.25pt}
        \end{tabular}
    }
    \vspace{-0.6em}
\end{table}

%% file: tables/ablation_extractor_parameters.tex
\begin{table}[t]
    \renewcommand\arraystretch{1.13}
    \caption{
    The parameters and computational overhead of signal extractors.
    }
    \label{tab:ablation_extractor_parameters}
    \centering
        \begin{tabular}{cccc}
            \Xhline{1.25pt}
            Model & Params. (M) & FLOPs (G) & Time (ms) \\
            \Xhline{0.75pt}
            HED & 14.04 & 160.42 & 2.96 \\
            DepthAnythingV2 & 93.27 & 311.81 & 	6.18 \\
            MaskDINO & 41.79 & 62.30 & 19.77 \\
            CLIP & 602.54 & 323.94 & 10.46 \\
            \Xhline{1.25pt}
        \end{tabular}
    \vspace{-0.2em}
\end{table}

%% file: tables/ablation_dpca.tex
\begin{table}[t]
    \renewcommand\arraystretch{1.2}
    \caption{
        Effectiveness of the DPCA.
    }
    \vspace{0.1em}
    \label{tab:ablation_dpca}
    \centering
    \resizebox{\linewidth}{!}{
        \begin{tabular}{ccccc}
        \Xhline{1.25pt}
        Methods & PSNR$\uparrow$&MANIQA$\uparrow$&MUSIQ$\uparrow$&CLIPIQA$\uparrow$\\
        \Xhline{0.75pt}
        Ours w/o DPCA&\textbf{23.6411}& 0.5371 & 69.3150 & 0.6553 \\
        \Xhline{0.75pt}
        \textbf{Ours}& 23.4909 &\textbf{0.5574}&\textbf{70.0252}&\textbf{0.6790}\\
        \Xhline{1.25pt}
        \end{tabular}
    }
    \vspace{-0.1em}
\end{table}

%% file: tables/ablation_lgwam.tex
\begin{table}[t]
    \renewcommand\arraystretch{1.4}
    \caption{
    Effectiveness of the LGWAM.
    }
    \label{tab:ablation_lgwam}
    \centering
    \resizebox{\linewidth}{!}{
        \begin{tabular}{ccccc}
            \Xhline{1.25pt}
            Methods & PSNR$\uparrow$ & MANIQA$\uparrow$ & MUSIQ$\uparrow$ & CLIPIQA$\uparrow$ \\
            \Xhline{0.75pt}
            Ours w/o LGWAM & 23.5731 & 0.5319 & 68.2780 & 0.6539 \\
            SeeSR w/ LGWAM & \textbf{23.6404} & 0.5272 & 68.5875 & 0.6720 \\
            \Xhline{0.75pt}
            \textbf{Ours} & 23.4909 & \textbf{0.5574} & \textbf{70.0252} & \textbf{0.6790} \\
            \Xhline{1.25pt}
        \end{tabular}
    }
    \vspace{-1.5em}
\end{table}

%% file: tables/ablation_signals.tex
\begin{table}[t]
    \renewcommand\arraystretch{1.2}
    \caption{
    Effectiveness of the signals.
    }
    \vspace{-0.4em}
    \label{tab:ablation_signals}
    \centering
    \resizebox{\linewidth}{!}{
        \begin{tabular}{ccccc}
            \Xhline{1.25pt}
            Methods & PSNR$\uparrow$ & MANIQA$\uparrow$ & MUSIQ$\uparrow$ & CLIPIQA$\uparrow$ \\
            \Xhline{0.75pt}
            Ours w/o (H \& D \& S) & \textbf{23.9221} & 0.5213 & 68.3985 & 0.6621 \\
            Ours w/o (D \& S) & 23.9197 & 0.5390 & 68.6778 & 0.6555 \\
            Ours w/o H & 23.8420 & 0.5291 & 68.9672 & 0.6606 \\
            Ours w/o S & 23.8978 & 0.5216 & 68.2412 & 0.6639 \\
            Ours w/o D & 23.7164 & 0.5336 & 68.5980 & 0.6653 \\
            \Xhline{0.75pt}
            \textbf{Ours} & 23.4909 & \textbf{0.5574} & \textbf{70.0252} & \textbf{0.6790} \\
            \Xhline{1.25pt}
        \end{tabular}
    }
\end{table}

%% file: tables/ablation_canny_sketch.tex
\begin{table}[t]
    \renewcommand\arraystretch{1.31}
    \caption{
    Signal sensitivity analysis.
    }
    \vspace{-0.5em}
    \label{tab:ablation_canny_sketch}
    \centering
        \begin{tabular}{ccccc}
            \Xhline{1.25pt}
            Methods & PSNR$\uparrow$ & MANIQA$\uparrow$ & MUSIQ$\uparrow$ & CLIPIQA$\uparrow$ \\
            \Xhline{0.75pt}
            Add Canny & \textbf{23.5729} & 0.5520 & 69.9131 & 0.6739 \\
            Add sketch & 23.4538 & 0.5484 & 69.4461 & \textbf{0.6821} \\
            \Xhline{0.75pt}
            \textbf{Ours} & 23.4909 & \textbf{0.5574} & \textbf{70.0252} & 0.6790 \\
            \Xhline{1.25pt}
        \end{tabular}
\end{table}

%% file: tables/overall_parameters.tex
\begin{table}[t]
    \renewcommand\arraystretch{1.28}
    \caption{Complexity analysis of different Real-ISR methods.}
    \vspace{-0.4em}
    \label{tab:overall_parameters}
    \centering
    \resizebox{\linewidth}{!}{
        \begin{tabular}{ccccc}
            \Xhline{1.25pt}
            Methods & Params. (M) & Train. Params. (M) & Runtime (s) & Steps \\
            \Xhline{0.75pt}
            BSRGAN & 16.70 & 16.70 & 0.08 & 1 \\
            Real-ESRGAN & 16.70 & 16.70 & 0.09 & 1 \\
            DASR & 8.07 & 8.07 & 0.05 & 1 \\
            \Xhline{0.75pt}
            StableSR & - & - & 3.52 & 50 \\
            DiffBIR & - & - & 6.87 & 50 \\
            SeeSR & 1615.8 & 749.9 & 4.34 & 50 \\
            HoliSDiP & 1627.8 & 761.2 & 5.72 & 50 \\
            \Xhline{0.75pt}
            MegaSR & 1669.6 & 803.7 & 6.35 & 50 \\
            \Xhline{1.25pt}
        \end{tabular}
    }
\end{table}

%% file: parts/VI_conclusion.tex
\section{Conclusion}
\label{sec:conclusion}

In this work, we first summarize and analyze the limitations of existing T2I-based Real-ISR methods through qualitative and quantitative evaluations.
Building upon the observations, we introduce MegaSR to mine fine-grained customized semantics and expressive guidance for real-world image super-resolution.
Specifically, we propose the Customized Semantics Module (CSM) to supplement fine-grained visual semantic knowledge and customize the proportions of multi-level semantics for different U-Net blocks.
Beyond semantic adaptation, we identify HED boundaries, depth maps, and semantic segmentation masks as crucial guidance signals through comparative experiments.
To this end, we first design prior-guided fine-tuning strategies to enhance the robustness of signal extractors for the real-world scenario.
Then, we design the Multimodal Signal Fusion Module (MSFM) to progressively incorporate multimodal guidance signals into the T2I model.
Extensive experiments on both real-world and synthetic datasets demonstrate the superiority of the proposed method in terms of semantic richness and structural consistency.

%% file: Bibliography.bib
@String(TPAMI = {IEEE TPAMI})

@String(TIP = {IEEE TIP})

@String(IJCV = {IJCV})

@String(ICCV = {ICCV})

@String(ICCVW = {ICCVW})

@String(CVPR = {CVPR})

@String(CVPRW = {CVPRW})

@String(ECCV = {ECCV})

@String(NIPS = {NeurIPS})

@String(ICLR = {ICLR})

@String(MICCAI = {MICCAI})

@String(ICML = {ICML})

@String(AAAI = {AAAI})

@String(ACMMM = {ACM MM})

@article{dong_srcnn_2016,
  author       = {Chao Dong and
                  Chen Change Loy and
                  Kaiming He and
                  Xiaoou Tang},
  title        = {Image Super-Resolution Using Deep Convolutional Networks},
  journal      = TPAMI,
  publisher    = {IEEE},
  pages        = {295--307},
  year         = {2016},
}

@article{zhong_dualsr_2025,
  title={Dual-Level Cross-Modality Neural Architecture Search for Guided Image Super-Resolution},
  author={Zhong, Zhiwei and Liu, Xianming and Jiang, Junjun and Zhao, Debin and Wang, Shiqi},
  journal=TPAMI,
  year={2025},
  publisher={IEEE}
}

@inproceedings{kim_vdsr_2016,
  author       = {Jiwon Kim and
                  Jung Kwon Lee and
                  Kyoung Mu Lee},
  title        = {Accurate Image Super-Resolution Using Very Deep Convolutional Networks},
  booktitle    = CVPR,
  pages        = {1646--1654},
  publisher    = {{IEEE}},
  year         = {2016},
}

@inproceedings{kim_drcn_2016,
  author       = {Jiwon Kim and
                  Jung Lee and
                  Kyoung Lee
                 },
  title        = {Deeply-Recursive Convolutional Network for Image Super-Resolution},
  booktitle    = CVPR,
  pages        = {1637--1645},
  publisher    = {{IEEE}},
  year         = {2016},
}

@inproceedings{tai_drrn_2017,
  author       = {Ying Tai and
                  Jian Yang and
                  Xiaoming Liu},
  title        = {Image Super-Resolution via Deep Recursive Residual Network},
  booktitle    = CVPR,
  pages        = {2790--2798},
  publisher    = {{IEEE}},
  year         = {2017},
}

@article{lai_fastlapsr_2018,
  title={Fast and accurate image super-resolution with deep laplacian pyramid networks},
  author={Lai, Wei-Sheng and Huang, Jia-Bin and Ahuja, Narendra and Yang, Ming-Hsuan},
  journal=TPAMI,
  pages={2599--2613},
  year={2018},
  publisher={IEEE}
}

@inproceedings{cheng_usr1_2019,
  author       = {Guoan Cheng and
                  Ai Matsune and
                  Qiuyu Li and
                  Leilei Zhu and
                  Huaijuan Zang and
                  Shu Zhan},
  title        = {Encoder-Decoder Residual Network for Real Super-Resolution},
  booktitle    = CVPRW,
  pages        = {2169--2178},
  publisher    = {{IEEE}},
  year         = {2019},
}

@inproceedings{zhang_rcan_2018,
  author       = {Yulun Zhang and
                  Kunpeng Li and
                  Kai Li and
                  Lichen Wang and
                  Bineng Zhong and
                  Yun Fu},
  title        = {Image Super-Resolution Using Very Deep Residual Channel Attention
                  Networks},
  booktitle    = ECCV,
  pages        = {294--310},
  publisher    = {Springer},
  year         = {2018},
}

@inproceedings{chen_ipt_2021,
  author       = {Hanting Chen and
                  Yunhe Wang and
                  Tianyu Guo and
                  Chang Xu and
                  Yiping Deng and
                  Zhenhua Liu and
                  Siwei Ma and
                  Chunjing Xu and
                  Chao Xu and
                  Wen Gao},
  title        = {Pre-Trained Image Processing Transformer},
  booktitle    = CVPR,
  pages        = {12299--12310},
  publisher    = {{IEEE}},
  year         = {2021},
}

@inproceedings{liang_swinir_2021,
  author       = {Jingyun Liang and
                  Jiezhang Cao and
                  Guolei Sun and
                  Kai Zhang and
                  Luc Van Gool and
                  Radu Timofte},
  title        = {SwinIR: Image Restoration Using Swin Transformer},
  booktitle    = ICCVW,
  pages        = {1833--1844},
  publisher    = {{IEEE}},
  year         = {2021},
}

@inproceedings{chen_dat_2023,
  author       = {Zheng Chen and
                  Yulun Zhang and
                  Jinjin Gu and
                  Linghe Kong and
                  Xiaokang Yang and
                  Fisher Yu},
  title        = {Dual Aggregation Transformer for Image Super-Resolution},
  booktitle    = ICCV,
  pages        = {12278--12287},
  publisher    = {{IEEE}},
  year         = {2023},
}

@article{chen_hat_2025,
  author       = {Xiangyu Chen and
                  Xintao Wang and
                  Wenlong Zhang and
                  Xiangtao Kong and
                  Yu Qiao and
                  Jiantao Zhou and
                  Chao Dong},
  title        = {{HAT:} Hybrid Attention Transformer for Image Restoration},
  journal      = TPAMI,
  publisher    = {IEEE},
  pages        = {1--18},
  year         = {2025},
}

@inproceedings{liu_catanet_2025,
  author       = {Xin Liu and
                  Jie Liu and
                  Jie Tang and
                  Gangshan Wu},
  title        = {CATANet: Efficient Content-Aware Token Aggregation for Lightweight
                  Image Super-Resolution},
  booktitle    = CVPR,
  pages        = {17902--17912},
  publisher    = {{IEEE}},
  year         = {2025},
}

@misc{goodfellow_gan_2014,
  author       = {Ian J. Goodfellow and
                  Jean Pouget{-}Abadie and
                  Mehdi Mirza and
                  Bing Xu and
                  David Warde{-}Farley and
                  Sherjil Ozair and
                  Aaron C. Courville and
                  Yoshua Bengio},
  title        = {Generative Adversarial Networks},
  note={arXiv:1406.2661},
  year={2014}
}

@inproceedings{ledig_srgan_2017,
  author       = {Christian Ledig and
                  Lucas Theis and
                  Ferenc Huszar and
                  Jose Caballero and
                  Andrew Cunningham and
                  Alejandro Acosta and
                  Andrew P. Aitken and
                  Alykhan Tejani and
                  Johannes Totz and
                  Zehan Wang and
                  Wenzhe Shi},
  title        = {Photo-Realistic Single Image Super-Resolution Using a Generative Adversarial
                  Network},
  booktitle    = CVPR,
  pages        = {105--114},
  publisher    = {{IEEE}},
  year         = {2017},
}

@article{hasan_pcsrgan_2025,
  author       = {Hasan, Md Rakibul and Behnoudfar, Pouria and MacKinlay, Dan and Poulet, Thomas},
  title        = {PC-SRGAN: Physically Consistent Super-Resolution Generative Adversarial Network for General Transient Simulations},
  journal      = TPAMI,
  publisher    = {IEEE},
  pages        = {12077--12083},
  year         = {2025},
}

@article{zhongxi_rethinkinggan_2023,
  title={Rethinking dual-stream super-resolution semantic learning in medical image segmentation},
  author={Qiu, Zhongxi and Hu, Yan and Chen, Xiaoshan and Zeng, Dan and Hu, Qingyong and Liu, Jiang},
  journal=TPAMI,
  pages={451--464},
  year={2023},
  publisher={IEEE}
}

@inproceedings{wang_realesrgan_2021,
  author       = {Xintao Wang and
                  Liangbin Xie and
                  Chao Dong and
                  Ying Shan},
  title        = {Real-ESRGAN: Training Real-World Blind Super-Resolution with Pure
                  Synthetic Data},
  booktitle    = ICCVW,
  pages        = {1905--1914},
  publisher    = {{IEEE}},
  year         = {2021},
}

@inproceedings{johnson_ploss_2016,
  author       = {Justin Johnson and
                  Alexandre Alahi and
                  Li Fei{-}Fei},
  title        = {Perceptual Losses for Real-Time Style Transfer and Super-Resolution},
  booktitle    = ECCV,
  pages        = {694--711},
  publisher    = {Springer},
  year         = {2016},
}

@inproceedings{simonyan_vgg_2015,
  author       = {Karen Simonyan and
                  Andrew Zisserman},
  title        = {Very Deep Convolutional Networks for Large-Scale Image Recognition},
  booktitle    = ICLR,
  publisher    = {OpenReview.net},
  year         = {2015},
}

@inproceedings{ronneberger_unet_2015,
  author       = {Olaf Ronneberger and
                  Philipp Fischer and
                  Thomas Brox},
  title        = {U-Net: Convolutional Networks for Biomedical Image Segmentation},
  booktitle    = MICCAI,
  pages        = {234--241},
  publisher    = {Springer},
  year         = {2015},
}

@inproceedings{miyato_ganinsta_2018,
  author       = {Takeru Miyato and
                  Toshiki Kataoka and
                  Masanori Koyama and
                  Yuichi Yoshida},
  title        = {Spectral Normalization for Generative Adversarial Networks},
  booktitle    = ICLR,
  publisher    = {OpenReview.net},
  year         = {2018},
}

@article{miaoyu_recdiffsr_2024,
  title={Latent diffusion enhanced rectangle transformer for hyperspectral image restoration},
  author={Li, Miaoyu and Fu, Ying and Zhang, Tao and Liu, Ji and Dou, Dejing and Yan, Chenggang and Zhang, Yulun},
  journal=TPAMI,
  year={2024},
  publisher={IEEE},
  pages={549--564},
}

@article{tong_stidiffsr_2024,
  title={Stimulating diffusion model for image denoising via adaptive embedding and ensembling},
  author={Li, Tong and Feng, Hansen and Wang, Lizhi and Zhu, Lin and Xiong, Zhiwei and Huang, Hua},
  journal=TPAMI,
  year={2024},
  publisher={IEEE},
  pages={8240--8257},
}

@inproceedings{yang_pasd_2024,
  author       = {Tao Yang and
                  Rongyuan Wu and
                  Peiran Ren and
                  Xuansong Xie and
                  Lei Zhang},
  title        = {Pixel-Aware Stable Diffusion for Realistic Image Super-Resolution
                  and Personalized Stylization},
  booktitle    = ECCV,
  pages        = {74--91},
  publisher    = {Springer},
  year         = {2024}
}

@inproceedings{wu_seesr_2024,
  author       = {Rongyuan Wu and
                  Tao Yang and
                  Lingchen Sun and
                  Zhengqiang Zhang and
                  Shuai Li and
                  Lei Zhang},
  title        = {SeeSR: Towards Semantics-Aware Real-World Image Super-Resolution},
  pages        = {25456--25467},
  publisher    = {{IEEE}},
  booktitle    = CVPR,
  year         = {2024}
}

@misc{xiao_segsr_2024,
  author       = {Jiahua Xiao and
                  Jiawei Zhang and
                  Dongqing Zou and
                  Xiaodan Zhang and
                  Jimmy S. J. Ren and
                  Xing Wei},
  title        = {Semantic Segmentation Prior for Diffusion-Based Real-World Super-Resolution},
  note={arXiv:2412.02960},
  year={2024}
}

@article{wang_stablesr_2024,
  author       = {Jianyi Wang and
                  Zongsheng Yue and
                  Shangchen Zhou and
                  Kelvin C. K. Chan and
                  Chen Change Loy},
  title        = {Exploiting Diffusion Prior for Real-World Image Super-Resolution},
  journal      = IJCV,
  pages        = {5929--5949},
  year         = {2024},
  publisher    = {Springer},
}

@inproceedings{lin_diffbir_2024,
  author       = {Xinqi Lin and
                  Jingwen He and
                  Ziyan Chen and
                  Zhaoyang Lyu and
                  Bo Dai and
                  Fanghua Yu and
                  Yu Qiao and
                  Wanli Ouyang and
                  Chao Dong},
  title        = {DiffBIR: Toward Blind Image Restoration with Generative Diffusion
                  Prior},
  booktitle    = ECCV,
  pages        = {430--448},
  publisher    = {Springer},
  year         = {2024},
}

@inproceedings{chen_faithdiff_2025,
  author       = {Junyang Chen and
                  Jinshan Pan and
                  Jiangxin Dong},
  title        = {FaithDiff: Unleashing Diffusion Priors for Faithful Image Super-resolution},
  booktitle    = CVPR,
  pages        = {28188--28197},
  publisher    = {{IEEE}},
  year         = {2025},
}

@inproceedings{he_resnet_2016,
  author       = {Kaiming He and
                  Xiangyu Zhang and
                  Shaoqing Ren and
                  Jian Sun},
  title        = {Deep Residual Learning for Image Recognition},
  booktitle    = CVPR,
  pages        = {770--778},
  publisher    = {{IEEE}},
  year         = {2016},
}

@inproceedings{li_blip_2022,
  author       = {Junnan Li and
                  Dongxu Li and
                  Caiming Xiong and
                  Steven C. H. Hoi},
  title        = {{BLIP:} Bootstrapping Language-Image Pre-training for Unified Vision-Language
                  Understanding and Generation},
  booktitle    = ICML,
  pages        = {12888--12900},
  publisher    = {{PMLR}},
  year         = {2022},
}

@misc{tsao_holisdip_2024,
  author       = {Li{-}Yuan Tsao and
                  Hao{-}Wei Chen and
                  Hao{-}Wei Chung and
                  Deqing Sun and
                  Chun{-}Yi Lee and
                  Kelvin C. K. Chan and
                  Ming{-}Hsuan Yang},
  title        = {HoliSDiP: Image Super-Resolution via Holistic Semantics and Diffusion
                  Prior},
  note={arXiv:2411.18662},
  year={2024}
}

@inproceedings{radford_clip_2021,
  author       = {Alec Radford and
                  Jong Wook Kim and
                  Chris Hallacy and
                  Aditya Ramesh and
                  Gabriel Goh and
                  Sandhini Agarwal and
                  Girish Sastry and
                  Amanda Askell and
                  Pamela Mishkin and
                  Jack Clark and
                  Gretchen Krueger and
                  Ilya Sutskever},
  title        = {Learning Transferable Visual Models From Natural Language Supervision},
  booktitle    = ICML,
  pages        = {8748--8763},
  publisher    = {{PMLR}},
  year         = {2021},
}

@inproceedings{rombach_ldm_2022,
  author       = {Robin Rombach and
                  Andreas Blattmann and
                  Dominik Lorenz and
                  Patrick Esser and
                  Bj{\"{o}}rn Ommer},
  title        = {High-Resolution Image Synthesis with Latent Diffusion Models},
  booktitle    = CVPR,
  pages        = {10674--10685},
  publisher    = {{IEEE}},
  year         = {2022},
}

@article{xie_hed_2017,
  author       = {Saining Xie and
                  Zhuowen Tu},
  title        = {Holistically-Nested Edge Detection},
  journal      = IJCV,
  pages        = {3--18},
  year         = {2017},
  publisher    = {Springer},
}

@inproceedings{yang_depthanythingv2_2024,
  author       = {Lihe Yang and
                  Bingyi Kang and
                  Zilong Huang and
                  Zhen Zhao and
                  Xiaogang Xu and
                  Jiashi Feng and
                  Hengshuang Zhao},
  title        = {Depth Anything {V2}},
  booktitle    = NIPS,
  year         = {2024},
  pages        = {21875--21911},
  publisher    = {NeurIPS Foundation},
}

@inproceedings{cheng_mask2former_2022,
  author       = {Bowen Cheng and
                  Ishan Misra and
                  Alexander G. Schwing and
                  Alexander Kirillov and
                  Rohit Girdhar},
  title        = {Masked-attention Mask Transformer for Universal Image Segmentation},
  booktitle    = CVPR,
  pages        = {1280--1289},
  publisher    = {{IEEE}},
  year         = {2022},
}

@inproceedings{li_maskdino_2023,
  author       = {Feng Li and
                  Hao Zhang and
                  Huaizhe Xu and
                  Shilong Liu and
                  Lei Zhang and
                  Lionel M. Ni and
                  Heung{-}Yeung Shum},
  title        = {Mask {DINO:} Towards {A} Unified Transformer-based Framework for Object
                  Detection and Segmentation},
  booktitle    = CVPR,
  pages        = {3041--3050},
  publisher    = {{IEEE}},
  year         = {2023},
}

@misc{zaremba_rnn_2014,
  author       = {Wojciech Zaremba and
                  Ilya Sutskever and
                  Oriol Vinyals},
  title        = {Recurrent Neural Network Regularization},
  note={arXiv:1409.2329},
  year={2014}
}

@inproceedings{vaswani_transformer_2017,
  author       = {Ashish Vaswani and
                  Noam Shazeer and
                  Niki Parmar and
                  Jakob Uszkoreit and
                  Llion Jones and
                  Aidan N. Gomez and
                  Lukasz Kaiser and
                  Illia Polosukhin},
  title        = {Attention is All you Need},
  booktitle    = NIPS,
  pages        = {5998--6008},
  year         = {2017},
  publisher    = {NeurIPS Foundation},
}

@inproceedings{li_lsdir_2023,
  author       = {Yawei Li and
                  Kai Zhang and
                  Jingyun Liang and
                  Jiezhang Cao and
                  Ce Liu and
                  Rui Gong and
                  Yulun Zhang and
                  Hao Tang and
                  Yun Liu and
                  Denis Demandolx and
                  Rakesh Ranjan and
                  Radu Timofte and
                  Luc Van Gool},
  title        = {{LSDIR:} {A} Large Scale Dataset for Image Restoration},
  booktitle    = CVPR,
  pages        = {1775--1787},
  publisher    = {{IEEE}},
  year         = {2023},
}

@inproceedings{bai_ffhq_2023,
  author       = {Haoran Bai and
                  Di Kang and
                  Haoxian Zhang and
                  Jinshan Pan and
                  Linchao Bao},
  title        = {{FFHQ-UV:} Normalized Facial UV-Texture Dataset for 3D Face Reconstruction},
  booktitle    = CVPR,
  pages        = {362--371},
  publisher    = {{IEEE}},
  year         = {2023},
}

@inproceedings{cai_realsr_2019,
  author       = {Jianrui Cai and
                  Hui Zeng and
                  Hongwei Yong and
                  Zisheng Cao and
                  Lei Zhang},
  title        = {Toward Real-World Single Image Super-Resolution: {A} New Benchmark
                  and a New Model},
  booktitle    = ICCV,
  pages        = {3086--3095},
  publisher    = {{IEEE}},
  year         = {2019},
}

@inproceedings{agustsson_div2kval_2017,
  author       = {Eirikur Agustsson and
                  Radu Timofte},
  title        = {{NTIRE} 2017 Challenge on Single Image Super-Resolution: Dataset and
                  Study},
  booktitle    = CVPRW,
  pages        = {1122--1131},
  publisher    = {{IEEE}},
  year         = {2017},
}

@inproceedings{wei_drealsr_2020,
  author       = {Pengxu Wei and
                  Ziwei Xie and
                  Hannan Lu and
                  Zongyuan Zhan and
                  Qixiang Ye and
                  Wangmeng Zuo and
                  Liang Lin},
  title        = {Component Divide-and-Conquer for Real-World Image Super-Resolution},
  booktitle    = ECCV,
  pages        = {101--117},
  publisher    = {Springer},
  year         = {2020},
}

@inproceedings{ignatov_dpediphone_2017,
  author       = {Andrey Ignatov and
                  Nikolay Kobyshev and
                  Radu Timofte and
                  Kenneth Vanhoey and
                  Luc Van Gool},
  title        = {DSLR-Quality Photos on Mobile Devices with Deep Convolutional Networks},
  booktitle    = ICCV,
  pages        = {3297--3305},
  publisher    = {{IEEE}},
  year         = {2017},
}

@inproceedings{ai_reallq250_2024,
  author       = {Yuang Ai and
                  Xiaoqiang Zhou and
                  Huaibo Huang and
                  Xiaotian Han and
                  Zhengyu Chen and
                  Quanzeng You and
                  Hongxia Yang},
  title        = {DreamClear: High-Capacity Real-World Image Restoration with Privacy-Safe
                  Dataset Curation},
  booktitle    = NIPS,
  year         = {2024},
  pages        = {55443--55469},
  publisher    = {NeurIPS Foundation},
}

@inproceedings{lim_df2k_2017,
  author       = {Bee Lim and
                  Sanghyun Son and
                  Heewon Kim and
                  Seungjun Nah and
                  Kyoung Mu Lee},
  title        = {Enhanced Deep Residual Networks for Single Image Super-Resolution},
  booktitle    = CVPRW,
  pages        = {1132--1140},
  publisher    = {{IEEE}},
  year         = {2017},
}

@inproceedings{Silberman_nyudepthv2_2012,
  author    = {Nathan Silberman, Derek Hoiem, Pushmeet Kohli and Rob Fergus},
  title     = {Indoor Segmentation and Support Inference from RGBD Images},
  booktitle = {ECCV},
  year      = {2012}
}

@article{russakovsky_imagenet1k_2015,
    Author = {Olga Russakovsky and Jia Deng and Hao Su and Jonathan Krause and Sanjeev Satheesh and Sean Ma and Zhiheng Huang and Andrej Karpathy and Aditya Khosla and Michael Bernstein and Alexander C. Berg and Li Fei-Fei},
    Title = { {ImageNet Large Scale Visual Recognition Challenge} },
    Year = {2015},
    journal   = IJCV,
    pages={211-252},
    publisher    = {Springer},
}

@inproceedings{alexander_ddpm_2021,
  title={Improved denoising diffusion probabilistic models},
  author={Nichol, Alexander Quinn and Dhariwal, Prafulla},
  booktitle=ICML,
  pages={8162--8171},
  year={2021},
  publisher    = {{PMLR}},
}

@inproceedings{william_dit_2023,
  title={Scalable diffusion models with transformers},
  author={Peebles, William and Xie, Saining},
  booktitle=ICCV,
  pages={4195--4205},
  year={2023},
  publisher    = {{IEEE}},
}

@article{wang_ssim_2004,
  author       = {Zhou Wang and
                  Alan C. Bovik and
                  Hamid R. Sheikh and
                  Eero P. Simoncelli},
  title        = {Image quality assessment: from error visibility to structural similarity},
  journal      = TIP,
  pages        = {600--612},
  year         = {2004},
  publisher    = {{IEEE}},
}

@inproceedings{zhang_lpips_2018,
  author       = {Richard Zhang and
                  Phillip Isola and
                  Alexei A. Efros and
                  Eli Shechtman and
                  Oliver Wang},
  title        = {The Unreasonable Effectiveness of Deep Features as a Perceptual Metric},
  booktitle    = CVPR,
  pages        = {586--595},
  publisher    = {{IEEE}},
  year         = {2018},
}

@article{zhang_niqe_2015,
  author       = {Lin Zhang and
                  Lei Zhang and
                  Alan C. Bovik},
  title        = {A Feature-Enriched Completely Blind Image Quality Evaluator},
  journal      = TIP,
  pages        = {2579--2591},
  year         = {2015},
  publisher    = {{IEEE}},
}

@inproceedings{yang_maniqa_2022,
  author       = {Sidi Yang and
                  Tianhe Wu and
                  Shuwei Shi and
                  Shanshan Lao and
                  Yuan Gong and
                  Mingdeng Cao and
                  Jiahao Wang and
                  Yujiu Yang},
  title        = {{MANIQA:} Multi-dimension Attention Network for No-Reference Image
                  Quality Assessment},
  booktitle    = CVPRW,
  pages        = {1190--1199},
  publisher    = {{IEEE}},
  year         = {2022},
}

@inproceedings{ke_musiq_2021,
  author       = {Junjie Ke and
                  Qifei Wang and
                  Yilin Wang and
                  Peyman Milanfar and
                  Feng Yang},
  title        = {{MUSIQ:} Multi-scale Image Quality Transformer},
  booktitle    = ICCV,
  pages        = {5128--5137},
  publisher    = {{IEEE}},
  year         = {2021},
}

@inproceedings{wang_clipiqa_2023,
  author       = {Jianyi Wang and
                  Kelvin C. K. Chan and
                  Chen Change Loy},
  title        = {Exploring {CLIP} for Assessing the Look and Feel of Images},
  booktitle    = AAAI,
  pages        = {2555--2563},
  publisher    = {{AAAI} Press},
  year         = {2023},
}

@misc{chen_pyiqa_2022,
  title={{IQA-PyTorch}: PyTorch Toolbox for Image Quality Assessment},
  author={Chaofeng Chen and Jiadi Mo},
  year={2022},
  howpublished = "[Online]. Available: \url{https://github.com/chaofengc/IQA-PyTorch}"
}

@inproceedings{zhang_bsrgan_2021,
  author       = {Kai Zhang and
                  Jingyun Liang and
                  Luc Van Gool and
                  Radu Timofte},
  title        = {Designing a Practical Degradation Model for Deep Blind Image Super-Resolution},
  booktitle    = ICCV,
  pages        = {4771--4780},
  publisher    = {{IEEE}},
  year         = {2021},
}

@inproceedings{liang_ldl_2022,
  author       = {Jie Liang and
                  Hui Zeng and
                  Lei Zhang},
  title        = {Details or Artifacts: {A} Locally Discriminative Learning Approach
                  to Realistic Image Super-Resolution},
  booktitle    = CVPR,
  pages        = {5647--5656},
  publisher    = {{IEEE}},
  year         = {2022},
}

@inproceedings{chen_femasr_2022,
  author       = {Chaofeng Chen and
                  Xinyu Shi and
                  Yipeng Qin and
                  Xiaoming Li and
                  Xiaoguang Han and
                  Tao Yang and
                  Shihui Guo},
  title        = {Real-World Blind Super-Resolution via Feature Matching with Implicit
                  High-Resolution Priors},
  booktitle    = ACMMM,
  pages        = {1329--1338},
  publisher    = {{ACM}},
  year         = {2022},
}

@inproceedings{liang_dasr_2022,
  author       = {Jie Liang and
                  Hui Zeng and
                  Lei Zhang},
  title        = {Efficient and Degradation-Adaptive Network for Real-World Image Super-Resolution},
  booktitle    = ECCV,
  pages        = {574--591},
  publisher    = {Springer},
  year         = {2022},
}

@inproceedings{wang_sinsr_2024,
  author       = {Yufei Wang and
                  Wenhan Yang and
                  Xinyuan Chen and
                  Yaohui Wang and
                  Lanqing Guo and
                  Lap{-}Pui Chau and
                  Ziwei Liu and
                  Yu Qiao and
                  Alex C. Kot and
                  Bihan Wen},
  title        = {SinSR: Diffusion-Based Image Super-Resolution in a Single Step},
  booktitle    = CVPR,
  pages        = {25796--25805},
  publisher    = {{IEEE}},
  year         = {2024},
}

@inproceedings{wu_osediff_2024,
  author       = {Rongyuan Wu and
                  Lingchen Sun and
                  Zhiyuan Ma and
                  Lei Zhang},
  title        = {One-Step Effective Diffusion Network for Real-World Image Super-Resolution},
  booktitle    = NIPS,
  pages        = {92529--92553},
  year         = {2024},
  publisher    = {NeurIPS Foundation},
}

@misc{bhat_zoedepth_2023,
  author       = {Shariq Farooq Bhat and
                  Reiner Birkl and
                  Diana Wofk and
                  Peter Wonka and
                  Matthias M{\"{u}}ller},
  title        = {ZoeDepth: Zero-shot Transfer by Combining Relative and Metric Depth},
  note={arXiv:2302.12288},
  year={2023}
}

@inproceedings{liu_swintransformerv2_2022,
  author       = {Ze Liu and
                  Han Hu and
                  Yutong Lin and
                  Zhuliang Yao and
                  Zhenda Xie and
                  Yixuan Wei and
                  Jia Ning and
                  Yue Cao and
                  Zheng Zhang and
                  Li Dong and
                  Furu Wei and
                  Baining Guo},
  title        = {Swin Transformer {V2:} Scaling Up Capacity and Resolution},
  booktitle    = CVPR,
  pages        = {11999--12009},
  publisher    = {{IEEE}},
  year         = {2022},
}

@inproceedings{zhang_ram_2024,
  author       = {Youcai Zhang and
                  Xinyu Huang and
                  Jinyu Ma and
                  Zhaoyang Li and
                  Zhaochuan Luo and
                  Yanchun Xie and
                  Yuzhuo Qin and
                  Tong Luo and
                  Yaqian Li and
                  Shilong Liu and
                  Yandong Guo and
                  Lei Zhang},
  title        = {Recognize Anything: {A} Strong Image Tagging Model},
  booktitle    = CVPRW,
  pages        = {1724--1732},
  publisher    = {{IEEE}},
  year         = {2024},
}

@misc{yang_qwen3_2025,
  author       = {An Yang and
                  Anfeng Li and
                  Baosong Yang and
                  Beichen Zhang and
                  Binyuan Hui and
                  Bo Zheng and
                  Bowen Yu and
                  Chang Gao and
                  Chengen Huang and
                  Chenxu Lv and
                  Chujie Zheng and
                  Dayiheng Liu and
                  Fan Zhou and
                  Fei Huang and
                  Feng Hu and
                  Hao Ge and
                  Haoran Wei and
                  Huan Lin and
                  Jialong Tang and
                  Jian Yang and
                  Jianhong Tu and
                  Jianwei Zhang and
                  Jian Yang and
                  Jiaxi Yang and
                  Jingren Zhou and
                  Junyang Lin and
                  Kai Dang and
                  Keqin Bao and
                  Kexin Yang and
                  Le Yu and
                  Lianghao Deng and
                  Mei Li and
                  Mingfeng Xue and
                  Mingze Li and
                  Pei Zhang and
                  Peng Wang and
                  Qin Zhu and
                  Rui Men and
                  Ruize Gao and
                  Shixuan Liu and
                  Shuang Luo and
                  Tianhao Li and
                  Tianyi Tang and
                  Wenbiao Yin and
                  Xingzhang Ren and
                  Xinyu Wang and
                  Xinyu Zhang and
                  Xuancheng Ren and
                  Yang Fan and
                  Yang Su and
                  Yichang Zhang and
                  Yinger Zhang and
                  Yu Wan and
                  Yuqiong Liu and
                  Zekun Wang and
                  Zeyu Cui and
                  Zhenru Zhang and
                  Zhipeng Zhou and
                  Zihan Qiu},
  title        = {Qwen3 Technical Report},
  note={arXiv:2505.09388},
  year={2025}
}

@misc{bai_qwen25vl_2025,
  author       = {Shuai Bai and
                  Keqin Chen and
                  Xuejing Liu and
                  Jialin Wang and
                  Wenbin Ge and
                  Sibo Song and
                  Kai Dang and
                  Peng Wang and
                  Shijie Wang and
                  Jun Tang and
                  Humen Zhong and
                  Yuanzhi Zhu and
                  Ming{-}Hsuan Yang and
                  Zhaohai Li and
                  Jianqiang Wan and
                  Pengfei Wang and
                  Wei Ding and
                  Zheren Fu and
                  Yiheng Xu and
                  Jiabo Ye and
                  Xi Zhang and
                  Tianbao Xie and
                  Zesen Cheng and
                  Hang Zhang and
                  Zhibo Yang and
                  Haiyang Xu and
                  Junyang Lin},
  title        = {Qwen2.5-VL Technical Report},
  note={arXiv:2502.13923},
  year={2025}
}

@inproceedings{zhao_unicontrolnet_2023,
  author       = {Shihao Zhao and
                  Dongdong Chen and
                  Yen{-}Chun Chen and
                  Jianmin Bao and
                  Shaozhe Hao and
                  Lu Yuan and
                  Kwan{-}Yee K. Wong},
  title        = {Uni-ControlNet: All-in-One Control to Text-to-Image Diffusion Models},
  booktitle    = NIPS,
  year         = {2023},
  pages        = {11127--11150},
  publisher    = {NeurIPS Foundation},
}

@inproceedings{hu_lora_2022,
  author       = {Edward J. Hu and
                  Yelong Shen and
                  Phillip Wallis and
                  Zeyuan Allen{-}Zhu and
                  Yuanzhi Li and
                  Shean Wang and
                  Lu Wang and
                  Weizhu Chen},
  title        = {LoRA: Low-Rank Adaptation of Large Language Models},
  booktitle    = ICLR,
  pages        = {3},
  publisher    = {OpenReview.net},
  year         = {2022},
}

@inproceedings{park_saft_2019,
  author       = {Taesung Park and
                  Ming{-}Yu Liu and
                  Ting{-}Chun Wang and
                  Jun{-}Yan Zhu},
  title        = {Semantic Image Synthesis With Spatially-Adaptive Normalization},
  booktitle    = CVPR,
  pages        = {2337--2346},
  publisher    = {{IEEE}},
  year         = {2019},
}

@inproceedings{lvmin_controlnet_2023,
  title={Adding conditional control to text-to-image diffusion models},
  author={Zhang, Lvmin and Rao, Anyi and Agrawala, Maneesh},
  booktitle=ICCV,
  pages={3836--3847},
  publisher    = {{IEEE}},
  year={2023}
}
